\documentclass[a4paper, 10pt, oneside, onecolumn]{elsarticle}

\makeatletter
\def\ps@pprintTitle{%
  \let\@oddhead\@empty
  \let\@evenhead\@empty
  \let\@oddfoot\@empty
  \let\@evenfoot\@empty}
\makeatother

\usepackage[utf8]{inputenc} 
\usepackage[T1]{fontenc}    
\usepackage{hyperref}       
\usepackage{url}            
\usepackage{booktabs}       
\usepackage{multirow}
\usepackage{amsfonts}       
\usepackage{nicefrac}       
\usepackage{microtype}      
\usepackage{wrapfig} 
\usepackage{xcolor}         
\usepackage{subcaption}
\usepackage{amsmath}
\usepackage{amsmath}
\usepackage{booktabs} 
\usepackage{makecell} 

\usepackage{url} 

\usepackage{amssymb}

\usepackage{lineno} 

\usepackage{amsmath}

\journal{Internet of Things}

\begin{document}

\begin{frontmatter}





\title{Exploiting Edge Features for Transferable Adversarial Attacks in Distributed Machine Learning}

\author[sssup]{Giulio Rossolini}
\ead{giulio.rossolini@santannapisa.it}
\author[unica]{Fabio Brau}
\ead{fabio.brau@unica.it}
\author[sssup]{Alessandro Biondi}
\ead{alessandro.biodni@santannapisa.it}
\author[unica]{\\Battista Biggio}
\ead{battista.biggio@unica.it}
\author[sssup]{Giorgio Buttazzo}
\ead{giorgio.buttazzo@santannapisa.it}

    \affiliation[sssup]{organization={Department of Excellence in Robotics and AI, TeCIP, Scuola Superiore Sant'Anna},
    addressline={Piazza Martiri della Libertà 33},
    postcode={56127},
    city={Pisa},
    country={Italy}}
\affiliation[unica]{organization={Department of Electrical and Electronic Engineering, University of Cagliari},
addressline={Via Marengo~2}, 
city={Cagliari},
postcode={09123}, 
country={Italy}}
\begin{abstract}
As machine learning models become increasingly deployed across the edge of internet of things environments, a partitioned deep learning paradigm in which models are split across multiple computational nodes introduces a new dimension of security risk.
Unlike traditional inference setups, these distributed pipelines span the model computation across heterogeneous nodes and communication layers, thereby exposing a broader attack surface to potential adversaries.
Building on these motivations, this work explores a previously overlooked vulnerability: even when both the edge and cloud components of the model are inaccessible (i.e., black-box), an adversary who intercepts the intermediate features transmitted between them can still pose a serious threat.
We demonstrate that, under these mild and realistic assumptions, an attacker can craft highly transferable proxy models, making the entire deep learning system significantly more vulnerable to evasion attacks. 
In particular, the intercepted features can be effectively analyzed and leveraged to distill surrogate models capable of crafting highly transferable adversarial examples against the target model.
To this end, we propose an exploitation strategy specifically designed for distributed settings, which involves reconstructing the original tensor shape from vectorized transmitted features using simple statistical analysis, and adapting surrogate architectures accordingly to enable effective feature distillation.

A comprehensive and systematic experimental evaluation has been conducted to demonstrate that surrogate models trained with the proposed strategy, i.e., leveraging intermediate features, tremendously improve the transferability of adversarial attacks. 
These findings underscore the urgent need to account for intermediate feature leakage in the design of secure distributed deep learning systems, particularly in edge scenarios, where constrained devices are more exposed to communication vulnerabilities and offer limited protection mechanisms.
\end{abstract}



\end{frontmatter}




\section{Introduction}
The growing adoption of deep learning across a wide range of application domains, including internet of things (IoT) and edge artificial intelligence (AI), has led to an increasing interest in partitioned learning and inference, paradigms in which a deep neural network (DNN) is divided across multiple heterogeneous nodes~\cite{verbraeken2020survey,Kang_collaborative,distributed_dnn_Teerapittayanon,Ko_partitioning,Anjos_survey_collaborative}. In a standard vertical partitioning scheme, the setting addressed in this work, the initial layers of the model are executed locally on the edge device, while the remaining layers are offloaded to the cloud. 

While the partitioned paradigm offers clear advantages in terms of efficiency and scalability, it also introduces new security vulnerabilities. In the broader context of secure and robust AI, deep learning models are known to be susceptible to evasion attacks~\cite{biggio2013evasion,szegedy2014intriguing}, in which adversaries craft subtle perturbations to input data that lead to incorrect predictions. Although a large part of the literature focuses on white-box scenarios, where the attacker has full access to model parameters and gradients, such assumptions are often unrealistic in practice. More practical threat models, in fact, adopt a black-box setting, in which the attacker has no knowledge of the architecture and inner parameters of the target model and can only rely on the outcome on known samples through a query-feedback mechanism. Under these constraints, a common procedure adopted by many black-box attacks involves first training a \emph{surrogate} model to approximate the \emph{target} model behavior and then crafting adversarial examples on this surrogate in the hope that they transfer to the target~\cite{papernot2017practical,lord2022attacking,tashiro2020diversity}.

\begin{figure}[ht]
    \centering
    \includegraphics[width=0.65\textwidth]{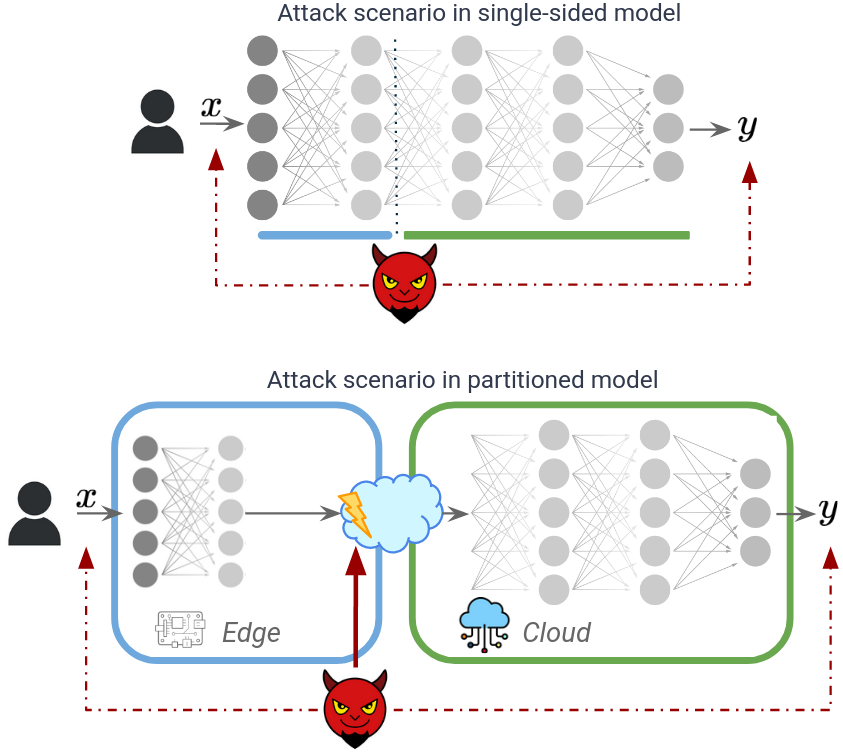}
    \caption{\small{Comparison of the classic black-box threat scenario with the proposed partitioned threat scenario. In our setting, an attacker can also exploit the transmitted intermediate features to build a highly transferable surrogate model.}}
    \label{fig:edge_partitioning}
\end{figure}

Under such a scenario, we argue that a partitioned inference introduces a broader and largely underexplored attack surface. In addition to inputs and outputs, partitioned models transmit intermediate features from the edge device to the cloud, which may be intercepted in unprotected or misconfigured environments. This represents a realistic threat in resource-constrained IoT deployments~\cite{guo2021robust_collaborative,security_imp_edge_comp} as further discussed in Section~\ref{s:motivations}. An adversary can leverage these intercepted features to train more transferable surrogate models, significantly enhancing the effectiveness of attacks, as illustrated in Figure~\ref{fig:edge_partitioning}.

\textbf{This work} investigates the vulnerabilities introduced in the partitioned model by revealing how intermediate feature leakage in partitioned inference can be exploited from an attacker’s perspective. A practical threat model is considered, in which an adversary is capable of intercepting serialized features transmitted between the edge and cloud. First, we demonstrate that the original tensor shape and spatial structure of these serialized features can reliably be reconstructed through statistical analysis. Building on this, we propose a feature-aware surrogate training strategy, which leverages the reconstructed hidden states to approximate the behavior of the target model more accurately.

An extensive experimental analysis has been conducted to evaluate the impact of exploiting intermediate features, highlighting the significant gains in adversarial transferability achieved by feature-aware training of the surrogate models. These improvements are consistently observed across different attack methods, resulting in substantially higher attack success rates. For example, in a query-based, i.e., black-box, attack such as GFCS-$\ell_{2}$~\cite{lord2022attacking} with \( \epsilon_{2} = 1.0 \), the success rate increases from 69\% to 96\% on a ResNet56 target when using a VGG16 surrogate trained with intermediate features, compared to a surrogate trained without access to them. Similarly, in the context of white-box attacks, a PGD-\( \ell_{\infty} \)\cite{pgd_attack} attack with \( \epsilon_{\infty} = \tfrac{8}{255} \) directed to the surrogate model achieves a 96\% success rate when transferred to the target model, whereas the success rate drops to only 61\% when intermediate features are unknown.

Supported by results presented in the following, this work offers, to the best of our knowledge, the first step toward the understanding of the effectiveness of feature distillation in enhancing surrogate model adversarial transferability. This is particularly relevant in practical scenarios involving partitioned neural networks, emphasizing the need to revisit existing security assumptions and to develop new defense mechanisms tailored to collaborative and distributed learning environments.


In summary, this work makes the following contributions:
\begin{itemize}
\item It presents a novel threat model for black-box attacks, which particularly fits with a partitioned scenario, emphasizing the exploitation of intermediate feature transmission between nodes.
\item It presents an efficient yet effective covariance matrix analysis technique that enables the reconstruction of the shape of feature tensors from their serialized form used for transmission between edge and cloud.
\item It proposes a feature distillation strategy within the context of black-box partitioned learning to build and train surrogate models, enabling attacks and principled security analysis based on the presented threat model.
\item It validates the proposed threat model and methodologies through an extensive suite of experiments, highlighting the significantly higher transferability of multiple attacks when using surrogate models informed by knowledge of internal features. Furthermore, an in-depth analysis is conducted by testing different partitioned configurations, e.g., varying the selection of split points, thereby providing useful insights and practical guidelines for the attacker.
\end{itemize}

The remainder of this paper is organized as follows.
Section \ref{s:motivations} reviews related work in security and partitioned learning, clarifying the motivations for our study.
Section \ref{s:motivations} formalizes the threat model and provides the necessary background.
Section \ref{s:method} describes the proposed methodology for reconstructing the shape of intercepted intermediate features and for training a feature-informed surrogate model.
Section \ref{s:exp} presents the experimental results.
Finally, Section \ref{s:conclusion} concludes the paper and outlines future research directions.

\section{Related work and motivations}
\label{s:motivations}

This section reviews the related work across two complementary perspectives. First, we examine the broader security challenges of distributed edge AI systems; then, we focus on evasion attacks and the role of surrogate models in enhancing black-box adversarial strategies.

\paragraph{Security challenges in distributed edge AI}
The evolution of distributed and collaborative learning paradigms has enabled the development of dedicated frameworks and studies~\cite{ray_framework_moritz,deepspeed_rasley}, allowing for reliable and efficient data and model sharing across multiple computational nodes. However, the distributed nature of these systems, which span heterogeneous devices and communication layers, significantly expands the attack surface, making them more susceptible to intrusions, unauthorized access, and data leakage~\cite{Sasi2024_iotsec_survey, Wei_IoTSec_survey, security_imp_edge_comp, He_collaborative_attacks_2021}.
For example, botnet attacks leveraged default credentials to compromise multiple of devices, while vulnerabilities in comunications stacks have enabled remote code execution and device compromise across diverse platforms~\cite{ZHANG2020_ioTsec_botnet_survey, Wei_IoTSec_survey}. Hardware-based side-channel attacks have also shown that attackers can exploit power, timing, or electromagnetic leakage from IoT devices to infer cryptographic keys or firmware contents \cite{SAYAKKARA2019_side_channel_survey}, which can then be used to extract transmitted packets between nodes.
These issues are especially pertinent in resource-constrained environments such as IoT and edge AI, where limited computational capabilities often coincide with weaker security protections. To better highlight the practical implications and feasibility of exploiting these weaknesses from an attacker point of view, Section \ref{ss:theat_model} presents a set of case studies that supports the threat model discussed.

All such issues suggest so that in AI-driven distributed systems intercepted data may include not only the inputs and outputs of DNNs, but also intermediate feature representations exchanged during inference. In such contexts, these vulnerabilities can expose the system to various AI attack strategies.
Previous research has begun to investigate these concerns in collaborative learning paradigms such as federated learning and split inference, with a primary focus on the integrity and confidentiality of AI models. For instance, studies have shown that intermediate representations can leak sensitive information~\cite{lyu2022privacy,liu2020backdoor_collaborative_attacks,He_collaborative_attacks_2021} and model authenticity can be compromised through backdoor insertion in distributed settings~\cite{Xie_distributed_backdoor_attacks_2020}.
However, while these efforts mainly addressed privacy leakage and training-time poisoning attacks, the impact of exposing intermediate features to evasion attacks (e.g., adversarial examples) is still not sufficiently explored~\cite{He_collaborative_attacks_2021, rossolini2024edge}.
Understanding and addressing such a vulnerability is crucial, as it reveals a realistic and novel threat model in partitioned inference systems.

\paragraph{Black-box attacks and the role of surrogate models}
Recent research has extensively demonstrated the vulnerability of DNNs to adversarial perturbations that lead to incorrect predictions~\cite{pgd_attack, CarliniAttack2017, goodfellow2014explaining, brau2022minimal}. A large portion of these studies focused on a white-box setting, where the attacker has full access to the target model’s architecture and parameters. In such cases, the attacker can directly compute gradients to optimize adversarial examples through well-established optimization techniques.
In contrast, a more realistic threat model (i.e., black-box) assumes that the attacker has no access to the weights and the internal structure of the target model. Here, the attacker can only interact with the system via input–output queries, which may reveal predicted class labels or, in some cases, full output probabilities~\cite{ilyas2018black, ilyas2018prior}.

Early black-box approaches were based on gradient-free optimization techniques, such as zeroth-order methods or evolutionary strategies~\cite{chen2017zoo, guo2019simple}, which approximate gradients by repeatedly querying the target model. However, these techniques often suffer from a prohibitively large number of queries required to mount an effective attack, which can comprise the feasibility and success of the attacks.
To mitigate this problem, recent approaches enhanced black-box attacks by leveraging surrogate models~\cite{tashiro2020diversity, cheng2019improving, lord2022attacking}. Such surrogates act as proxies for the target and guide the gradient estimation process, significantly reducing, or even eliminating, the need for queries and improving the success of the attack. In this context, it is worth noting that the effectiveness of such methods largely depends on the transferability property between the surrogate and the target model; that is, adversarial examples crafted to mislead the surrogate are likely to transfer to the target~\cite{demontis2019adversarial}.

Despite their effectiveness, many existing works assume overly optimistic conditions for constructing surrogate models, often relying on the public availability of pretrained models that closely resemble the target \cite{gu2023survey_atrans, zhu2024_survey_atrans_black_box}. This assumption rarely holds in practical applications, particularly in industrial or IoT scenarios, where the attacker cannot depend on publicly available architectures or datasets \cite{CAI_NEURIPS2022_surrogate, Qin_Xiong_Yi_Hsieh_2023_AAAI_surrogate}. Papernot et al.~\cite{papernot2017practical} addressed this challenge by proposing a method to train surrogate models through knowledge distillation based on the target input-output behavior using offline queries, a concept that also underpins techniques in model extraction and model stealing \cite{tramer2016_stealing, oliynyk2023know_stealing}.

Given the importance of transferability of the surrogate model in black-box attacks, yet often limited by restricted access in real-world scenarios, this work argues that, in partitioned AI systems, intercepted intermediate features offer a powerful and underexplored source of knowledge to boost surrogate fidelity beyond input–output distillation.

\section{Background and threat model}
This section introduces the necessary formalism for discussing the methodologies used in partitioned DNNs scenarios and evasion attacks. Then, it formalizes important properties and assumptions of the proposed threat model.

\subsection{Classic and partitioned DNNs}
In a standard (non-partitioned) model setting, an input sample \( x \in \mathbb{R}^{N_x} \) is processed by a DNN \( f: \mathbb{R}^{N_x} \rightarrow \mathbb{R}^{N_y} \) to produce an output \( y = f(x) \). Here, \( N_x \) and \( N_y \) denote the dimensionalities of the input and output spaces, respectively. For instance, in the case of classical image classification, \( N_x \) typically represents the input tensor shape \( C_x \times H_x \times W_x \) (i.e., number of input channels, height, and width), while \( N_y \) corresponds to the number of output classes. This classical DNN setup assumes full model availability and no splitting between multiple computing nodes.

In contrast, \textit{partitioned inference} (also referred to as split inference) introduces a structural decomposition of the model across different execution environments, typically involving an \textit{edge} device and a \textit{cloud} server. For simplicity, we consider a bipartite split into these two components; however, the notation and methodology can be readily extended to settings with more than two partitioned nodes.

As illustrated in Figure~\ref{fig:edge_partitioning}, the target model \( f \) is decomposed into two sequential components. The \textit{edge-side model}, denoted by \( f_{\text{e}}: \mathbb{R}^{N_x} \rightarrow \mathbb{R}^{N_{f_e}} \), is deployed on an edge node and consists of \( L_e \) layers that transform the input \( x \) into an intermediate representation of dimension \( N_{f_e} \). The \textit{cloud-side model}, denoted by \( f_{\text{c}}: \mathbb{R}^{N_{f_e}} \rightarrow \mathbb{R}^{N_y} \) is deployed in the cloud and comprises the remaining \( L_c \) layers that map the intermediate features to the final output prediction.
The overall model is thus defined as the composition:
\[
f(x) = (f_{\text{c}} \circ f_{\text{e}})(x) = f_{\text{c}}(f_{\text{e}}(x)),\quad \forall x\in\mathbb{R}^{N_x},
\]
\noindent where the \( \circ \) operator denotes the composition of functions.

\subsection{Adversarial attacks and surrogates}

Adversarial attacks aim at crafting small, human-imperceptible perturbations \(\delta\) applied to an input \(x \) with label \(y\), such that the perturbed input \(\tilde{x} = x + \delta\) causes the target model \(f\) to produce a wrong classification, i.e., \(f(\tilde{x}) \neq y\). In the case of \textit{untargeted attacks}, the objective is simply to induce any incorrect prediction, and the problem is typically formulated as the following optimization:
\begin{equation}
\min_{\delta} \, \mathcal{L}_{\textit{Adv}}(f(x + \delta), y) \quad \text{subject to} \quad \|\delta\|_p \leq \epsilon,
\label{eq:opt_adv}
\end{equation}
where \(\mathcal{L}_{\textit{Adv}}\) is a classification loss function (e.g., opposite of the cross-entropy), \(\epsilon\) bounds the perturbation magnitude, and \(p\) denotes the chosen norm (commonly \(p = 2\) or \(p = \infty\)).
In practice, in addition to constraining the perturbation magnitude, a box constraint is also applied during optimization to ensure that the final input remains in a valid format; for example, pixel values are constrained to lie within the range \([0, 1]\).

Solving the problem of Equation~\eqref{eq:opt_adv} primarily requires a white-box setting, where the parameters and gradients of the target model \( f \) are fully accessible to the attacker. However, this assumption does not hold in more constrained scenarios, as discussed in Section~\ref{s:motivations}, where only limited information about \( f \), such as input–output query pairs, can be leveraged. In such cases, the attacker operates in a black-box setting, where the use of a proxy surrogate model becomes essential. For example, white-box attack techniques can be applied to optimize adversarial examples on a surrogate model, which may then be used to attack target models, assuming sufficient transferability~\cite{demontis2019adversarial}.
This surrogate, denoted as \( g: \mathbb{R}^{N_x} \rightarrow \mathbb{R}^{N_y} \), must be trained in advance to approximate the behavior of \( f \), assuming that no pretrained version is available for the task at hand.\medskip

\noindent\textbf{Surrogate distillation.}
A standard approach for training such a surrogate model under the black-box setting is to minimize a \textit{similarity loss} between the surrogate’s output and that of the target model, computed over a set of queries based on samples available to the attacker \( \mathcal{X} = \{x_i, ..., x_N\} \). When the predicted probabilities or logits of \( f \) are accessible, the similarity loss can be defined as:

\begin{equation}
\mathcal{L}_{\text{out}} = \frac{1}{N} \sum_{i=1}^{N} \mathcal{D}\left(g(x_i), f(x_i)\right),
\label{eq:surrogate_loss}
\end{equation}

\noindent where \( \mathcal{D}(\cdot, \cdot) \) is a divergence or distance function, such as the Kullback–Leibler divergence or the mean squared error \cite{hinton2015distilling, xu2024survey_distill}.

In this context, ensuring that the surrogate model closely replicates the behavior of the black-box target model is crucial for improving attack transferability. We argue that significant insights can be gained by extending this framework to incorporate the knowledge of intermediate features of the target model, when such information is available. This additional supervision can further enhance the training of the surrogate model and improve the effectiveness of Equation~\eqref{eq:surrogate_loss} in aligning the surrogate's internal representations with those of the target.

\subsection{Threat model}
\label{ss:theat_model}

Inspired by the relevance of black-box attacks in classical DNN settings, this paper investigates a more relaxed yet practical black-box threat scenario that is particularly relevant for distributed learning and inference. In this setting, the target model \( f \) is partitioned across two nodes, such that \( f = f_{\text{c}} \circ f_{\text{e}} \). A key assumption in such a threat model is that the attacker, in addition to potentially observing the final outputs of \( f \), can also intercept the intermediate features \( f_{\text{e}}(x) \) transmitted between the edge and cloud nodes, while still not having access to the model parameters of the edge part $f_{\text{e}}$. This reflects a realistic attack surface in collaborative inference scenarios, where communication links may be vulnerable to passive inspection (e.g., side-channel attacks or data sniffing), as further discussed in the related work (Section~\ref{s:motivations}) and in some practical exploitation cases below.
Under this threat model, the next section shows how the attacker can train a surrogate model \( g \), which is also partitioned as \( g = g_{\text{c}} \circ g_{\text{e}} \), with the objective of approximating both the intermediate representations and the final output behavior of the target model.
To achieve this, the surrogate model is trained on the attacker dataset \( \mathcal{X} \) using multiple-term objective function that combines feature-level and output-level supervision.\medskip

\noindent \textbf{Main challenges.} Although the adopted threat model represents a relaxed variant of the classical black-box threat model, it is highly relevant in the context of distributed learning. 
Nevertheless, the practical realization of this threat model involves several non-trivial challenges, even under the assumption that intermediate features can be intercepted during transmission.

\begin{enumerate}
    \item \textit{Unknown feature shape due to serialization}. Even when the attacker is able to intercept the transmitted features, the raw tensor format is often serialized for transmission (e.g., flattened for compatibility with communication protocols). As a result, the original dimensionality \( N_{f_e}= (C_e, H_e, W_e) \) of the feature tensor is no longer explicitly available and is typically reconstructed only on the cloud side.

    \item \textit{Unknown architecture and splitting point}. Since the architecture of the target model \( f \) is unknown, and the exact location of the target splitting point is also unknown, it is infeasible to directly identify a surrogate architecture that contains an intermediate layer matching the dimensionality of the intercepted features.
\end{enumerate}

\noindent \textbf{Practical exploitations.}
This section presents two cases of exploitation that are practically relevant under the adopted threat model.

\emph{Case 1)} Edge and cloud nodes communicate with encrypted communication. Model inference on the edge node is handled in a \emph{trusted execution environment} (TEE) so that unauthorized infiltration in the edge node at the application and operating system level does not allow obtaining access to the model parameters. The attacker can however dispose of a side channel, e.g., stimulated by transient execution~\cite{canella2019transient}, to exfiltrate a small portion of information from the TEE, thereby being capable to retrieve the intermediate features \( f_{\text{e}}(x) \) or the key used to encrypt data in the edge-cloud communication, while not being capable to exfiltrate all model parameters due to the very low bandwidth of the side channel.

\emph{Case 2)} The edge and cloud parts of the model are made available \emph{as a service} from a third party~\cite{Ding2020service}. The attacker is a service user and intends to craft adversarial attacks for the split model with the purpose of damaging other service users. Being a service user, the attacker has legitimate access to the intermediate features but not to the model parameters.

\section{Exploiting the vulnerabilities}
\label{s:method}

To address the first challenge and reconstruct the original feature shape from the attacker's point of view, Section~\ref{s:shape_analysis} presents an efficient yet effective method based on covariance matrix analysis. By collecting a batch of intercepted feature transmissions, it is possible to estimate the spatial structure of the features, enabling a reliable reconstruction of the intermediate feature tensor.
The second challenge is addressed in Section~\ref{s:method_surrogate} by introducing an adaptation block that allows selecting a compatible splitting point in the surrogate architecture. This adaptation block transforms the surrogate's intermediate tensors to match the dimensionality of the intercepted features from the target model. As a result, it enables effective knowledge transfer through the distillation loss in Equation~\eqref{eq:surrogate_loss}.

\subsection{Estimating feature spatial dimensions}
\label{s:shape_analysis}

As commonly studied in AI models for vision tasks, the feature tensor produced by the edge-side model \( f_e(x) \) is typically structured as \( \mathcal{N}_{f_e} = (C_{f_e}, H_{f_e}, W_{f_e}) \), with \( C_{f_e} \), \( H_{f_e} \), and \( W_{f_e} \) denoting the number of channels, height, and width of the edge features, respectively. The batch dimension is omitted for simplicity (e.g., we work under the assumption that attackers can observe outputs on a per-sample basis).

In partitioned inference settings, intermediate feature tensors are transmitted over the network in a vectorized (i.e., flattened) form, thereby obscuring their original spatial and channel-wise structure. In accordance with the proposed threat model, during client-to-server transmission, this flattened representation is accessible to a potential adversary, while internal features in the edge and cloud parts are unknown. We denote this as a vector \( \vec{h} = \text{seq}(f_e(x)) \), where \( \vec{h} \in \mathbb{R}^d \) and \( d = C_{f_e} \cdot H_{f_e} \cdot W_{f_e} \). 
From the attacker's perspective, the key challenge is to recover the original spatial dimensions \( (C_{f_e}, H_{f_e}, W_{f_e}) \) from  \( \vec{h}  \). 

As a first step, we assume that the input images, and, consequently, the intermediate feature maps, are square-shaped (\(H_{f_e} = W_{f_e}\)), a common design choice in many deep learning frameworks. Under this assumption, it is shown that it is possible to estimate the spatial resolution, without prior knowledge of the number of channels \( C_{f_e} \), by analyzing patterns in the statistical distribution of \( \vec{h} \), such as periodicity or autocorrelation.

Specifically, let \( \mathcal{B} = \{x_1, \dots, x_N\} \) be a batch of \( N \) input images, and let \( \{\vec{h}_1, \dots, \vec{h}_N\} \subset \mathbb{R}^d \) be the corresponding set of flattened feature vectors extracted from the edge model. We construct a feature matrix \( X \in \mathbb{R}^{N \times d} \), where each row corresponds to a flattened feature vector:
\( X = [\vec{h}_1^\top; \vec{h}_2^\top; \dots; \vec{h}_N^\top] \).
We then compute the sample covariance matrix:
\begin{subequations}
\begin{align}
\Sigma &= \frac{1}{N} (X - \bar{X})^\top (X - \bar{X}) \in \mathbb{R}^{d \times d} \\
\text{where } \bar{X} &= \frac{1}{N} \sum_{i=1}^N \vec{h}_i\in\mathbb{R}^d.
\end{align}
\label{eq:covariance}
\end{subequations}

\begin{figure}[ht]
    \centering
    \includegraphics[width=0.9\textwidth]{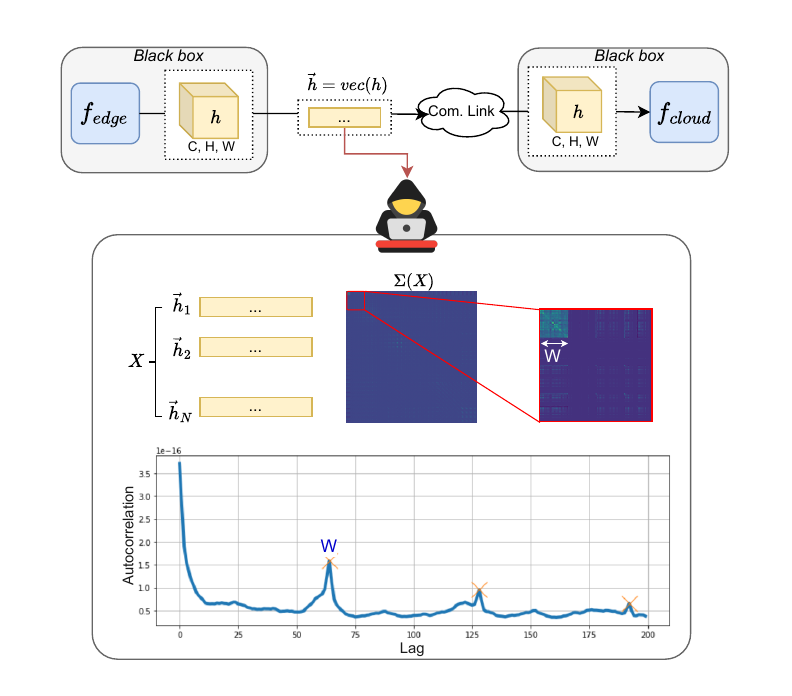}
    \caption{\small{Illustration of the proposed shape reconstruction approach based on covariance matrix analysis of serialized feature vectors intercepted by the attacker: the attacker sniff and store a batch of flattened feature vectors and computes their covariance matrix as described in Equation~\eqref{eq:covariance}. The resulting matrix reveals spatial patterns that, by construction, reflect the original feature map’s shape. In particular, the width \( W_{f_e} \) (i.e., the last spatial dimension of the unknown feature tensor) can be estimated automatically through a peak detection analysis of the autocorrelation, as shown in the bottom plot of the figure, where the lag $k$ is reported on the x-axis and the autocorrelation $R(k)$ on the y-axis.
}}
    \label{fig:correlation}
\end{figure}

As illustrated in Figure~\ref{fig:correlation}, the matrix \( \Sigma \) reveals a grid-like structure that reflects the spatial regularity of the original feature maps. In particular, periodic block diagonals emerge, corresponding to correlations between spatially adjacent rows in the original two-dimensional layout. This results in visually distinct square-shaped blocks of size \( W_{f_e} \times W_{f_e} \) along the main diagonal of \( \Sigma \). 
By analyzing the number and size of these blocks, either through visual inspection or via automated methods, such as peak detection, we can accurately estimate the spatial width \( W_{f_e} \). 
To automatically analyze matrix \( \Sigma \), it is possible to compute 
the autocorrelation of row-wise mean as $R(k) = \sum_{i=1}^{N-k} \mu_i \cdot \mu_{i+k}$, 
where \( \mu_i \) denotes the mean of the \( i \)-th row of \( \Sigma \), and \( k \geq 0 \) is the \emph{lag} value, representing how far the row-mean vector is shifted to evaluate its similarity with itself (i.e., to estimate possible patch sizes). The size $W_{f_e}$ can then be estimated by finding the value of $k$ for which $R(k)$ exhibits its highest peak—i.e., a point where the function ascends and then descends.
An example of autocorrelation analysis is illustrated at the bottom of 
Figure~\ref{fig:correlation} for $k \in [0, 200]$ and $N = 256$.

In the experimental section, the effectiveness of the proposed approach is validated under a variety of conditions, evaluating the impact of batch size and testing across different network architectures as well as various edge/cloud partitioning points.\medskip

\noindent\textbf{Considerations on non-square dimensions.}
As stated in the previous section, many deep learning models are designed with square input resolutions (i.e., \( H_{x} = W_{x} \)), particularly in standard benchmark datasets. However, this assumption does not always hold in some domains, where rectangular input dimensions are common~\cite{cordts2016cityscapes}. In such cases, estimating only the spatial width \( W_{f_e} \) leads to an underdetermined expression for the product \( H_{f_e} \cdot C_{f_e} = d / W_{f_e} \), which entangles the unknown height \( H_{f_e} \) with the unknown number of channels \( C_{f_e} \).
Nevertheless, if the attacker has access to the resolution of the original input images, this provides knowledge of the aspect ratio between spatial dimensions, defined as \( r = H_{x} / W_{x} \), which is generally preserved across the convolutional layers in CNN architectures. Given this, once the spatial width \( \hat{W}_{f_e} \) is estimated using the proposed method, the corresponding height can be recovered as
\(
\hat{H}_{f_e} = r \cdot \hat{W}_{f_e}.
\)

\subsection{Building the partitioned surrogate model}
\label{s:method_surrogate}
Once the original shape of the intercepted intermediate features is extracted, we can design and train an effective surrogate model to produce highly transferable adversarial attacks.\medskip 

\noindent\textbf{Adapting a reference surrogate model.}
In the considered scenario, it is essential to define a surrogate model \( g = g_c \circ g_e \), where the edge component \( g_e \) produces features whose shape \( \mathcal{N}_{g_e} \) matches the intercepted feature shape \( \mathcal{N}_{f_e} \) of the target edge model \( f_e \).
To achieve this, we start from a reference surrogate architecture, denoted by \( \bar{g} \), and select an internal split layer such that \( \bar{g} = \bar{g}_c \circ \bar{g}_e \). At this initial stage, the output features of \( \bar{g}_e \) may differ from those of the intercepted features \( f_e(x) \), both in spatial resolution and channel dimensionality.\footnote{Since the architecture of the target model is unknown, there may not exist a layer in the surrogate architecture that exactly matches the desired spatial and channel dimensions. Therefore, it is necessary to adapt the surrogate accordingly by modifying its structure.}

\begin{figure}[ht]
    \centering
    \includegraphics[width=1.0\textwidth]{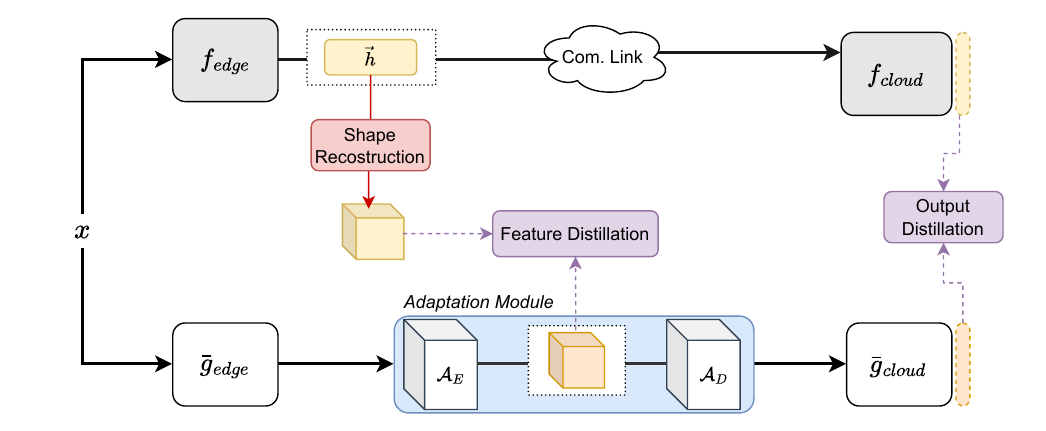}
    \caption{\small{Illustration of the proposed adaptation procedure, where the final surrogate model \( g \) integrates the original edge and cloud components of the reference surrogate model \( \bar{g} \) along with an \textit{adaptation module} that enables matching the intermediate features intercepted from the unknown target model and performing feature distillation.
    In the figure, subscripts `e' and `c' are expanded into `edge' and `cloud', respectively.
}}
    \label{fig:aux_block}
\end{figure}

Such a mismatch was resolved by introducing an \emph{adaptation encoder block} \( \mathcal{A}_E \), placed immediately after \( \bar{g}_e \), which transforms its output to match the dimensions of the intercepted features \( f_e(x) \). The resulting composition defines the final edge-side component of the surrogate as \( g_e = \mathcal{A}_E \circ \bar{g}_e \), as illustrated in Figure~\ref{fig:aux_block}.
The design of the adaptation block is straightforward and addresses mismatches in both the channel and spatial dimensions. For channel alignment, a \(1 \times 1\) convolutional layer maps the original channel dimension \( C_{\bar{g}_e} \) to the target \( C_{f_e} \). The spatial mismatch is addressed by employing a cascade of trainable upscaling and downscaling modules, implemented using transposed convolution and standard convolution layers, respectively. These are followed by an interpolation layer to precisely match the spatial dimensions \( (H_{f_e}, W_{f_e}) \) of the intercepted feature tensor.

Once the surrogate edge component \( g_e \) is aligned with the target edge output, the input is also adapted to the original surrogate cloud component \( \bar{g}_c \). This is done by using an \emph{adaptation decoder block} \( \mathcal{A}_D \), which consists of a convolutional layer and interpolation to reshape the features into the expected dimensions by \( \bar{g}_c \), i.e., \( (H_{\bar{g}_e}, W_{\bar{g}_e}) \).

The final surrogate model is shown at the bottom of Figure~\ref{fig:aux_block}, where the \textit{adaptation module} includes both the encoder \( \mathcal{A}_E \) and decoder \( \mathcal{A}_D \) blocks described above. The edge and cloud parts of the final surrogate model can be defined as:
\begin{equation}
\begin{aligned}
g_e &= \mathcal{A}_E \circ \bar{g}_e, \\
g_c &= \bar{g}_c \circ \mathcal{A}_D.
\label{eq:aux_block}
\end{aligned}
\end{equation}

Please note that the selection of the split point in the original surrogate architecture is not straightforward. This is because the attacker lacks information about the depth at which the model is partitioned in the target architecture. As a result, a careful analysis is required to select an appropriate split layer in the surrogate. This challenge is further explored and discussed in Section \ref{s:exp}. \medskip

\noindent\textbf{Training the surrogate model.}
The surrogate model \( g = g_{\text{c}} \circ g_{\text{e}} \) is trained using two complementary loss functions that exploit the available information: the intermediate features \( f_{\text{e}}(x) \) from the target edge model and the final output prediction \( f(x) \) from the complete target model.

In particular, the surrogate parameters \( \Theta_g \) are learned by minimizing the following loss:
\(
\Theta_g = \arg\min_{\Theta_g} \sum_{x \in \mathcal{X}} \mathcal{L}(g, f, x),
\)
where the combined loss function is defined as:
\begin{equation}
\mathcal{L}(g, f, x) = \alpha \cdot \mathcal{L}_{\text{features}}(g_{\text{e}}, f_{\text{e}}, x) + \beta \cdot \mathcal{L}_{\text{out}}(g, f, x).
\end{equation}
The first term, \( \mathcal{L}_{\text{features}} \), encourages the surrogate's edge model \( g_{\text{e}} \) to learn features that closely resemble those extracted by the target edge model \( f_{\text{e}} \). The second term, \( \mathcal{L}_{\text{out}} \), supervises the surrogate’s cloud model \( g_{\text{c}} \), and is defined according to the type of output information accessible from the target model. The coefficients \( \alpha, \beta \in \mathbb{R}_{\geq 0} \) are scalar weights that balance the contributions of feature-level and output-level distillation.

For the feature distillation term, $\mathcal{L}_{\text{features}}$, we adopt the mean squared error (MSE), defined as \( \| f_{\text{e}}(x) - g_{\text{e}}(x) \|_2^2 \), which is commonly used in feature distillation tasks~\cite{heo2019comprehensive, xu2024survey_distill}.

For the output loss \( \mathcal{L}_{\text{out}} \), we adapt the formulation based on the type of output information available from the target model. If the full output probability vector is accessible, we use the Kullback–Leibler divergence:
\(
\mathcal{L}_{\text{out}}(g, f, x)  = \sum_{i} f_i(x) \log \left( \frac{f_i(x)}{g_i(x)} \right).
\)
If only the predicted class \( \hat{y} = \arg\max f(x) \) is available, we employ the cross-entropy loss:
\(
\mathcal{L}_{\text{out}}(g, f, x) = \mathcal{L}_{\text{CE}}(g(x), \hat{y}).
\)
Finally, if no output from the target is observable but the ground-truth label \( y \) is known, we fall back to the standard supervised setting:
\(
\mathcal{L}_{\text{out}}(g, f, x) = \mathcal{L}_{\text{CE}}(g(x), y).
\)
In the experimental section, we evaluate the performance of the surrogate model across all these settings.

\section{Experiments}\label{s:exp}
In the first part of the experiments (Section~\ref{ss:est_edge_shape}), we evaluate the ability to promptly and accurately estimate the shape of the intermediate features generated at the edge, leveraging the covariance-based analysis introduced in Section~\ref{s:shape_analysis}. In the second part of the experimental evaluation, we investigate the effect of feature distillation on training the surrogate model.
To this end, we compare the effectiveness of several state-of-the-art attack methods, detailed within each evaluation setting in the following sections, using the surrogate models trained with target-feature distillation with respect to classic output-only distillation. 


\subsection{Experimental setup}
The experiments focus on the CIFAR-10 dataset \cite{cifar10}, as it is commonly used in collaborative inference benchmarks~\cite{rossolini2024edge, matsubara2022split} and allows for evaluating multiple model training scenarios. We employ a class-balanced random split of the CIFAR-10 test set: half (5000 samples) is used to train the surrogate model, and the remaining half is used to evaluate the success of the attacks. This separation ensures that the surrogate model is trained on data distinct from that used to train the target models, thereby preserving the integrity of testing time distillation.
For training the surrogate model, we set the learning rate of the Adam optimizer to 0.05, along with its default parameters, and the distillation weights as \(\alpha\) and \(\beta\) are set to 0.5
We consider different CNN architectures as target models, commonly used in split inference~\cite{rossolini2024edge, matsubara2022split}, including VGG16~\cite{vgg_simonyan2014very}, ResNet56~\cite{he2016deep}, and MobileNetV2~\cite{howard2017mobilenets}\footnote{We use pretrained versions of these models on CIFAR-10, available at \url{https://github.com/huyvnphan/PyTorch_CIFAR10}}.

As a reference model for crafting adversarial examples, we adopt VGG16 trained from scratch under multiple settings for comparison. For the main analysis reported in Sections \ref{ss:query_based} and \ref{ss:white_box}, we focus on an intermediate splitting point, specifically, the second block, for both surrogate and target models. This choice reflects a typical partitioning observed in partitioned inference~\cite{rossolini2024edge, matsubara2022split}, where early layers up to approximately the mid-depth of the model are executed on the edge device. Furthermore, in Section~\ref{ss:split_depth} and Section \ref{ss:est_edge_shape}, we present an in-depth evaluation of how the relative positioning of split points, both in the surrogate and target models, affects attack effectiveness and allows extraction of the feature shapes.




\subsection{Estimation of edge feature shapes}
\label{ss:est_edge_shape}
\begin{figure}[!h]
    \centering
    \includegraphics[width=0.6\textwidth]{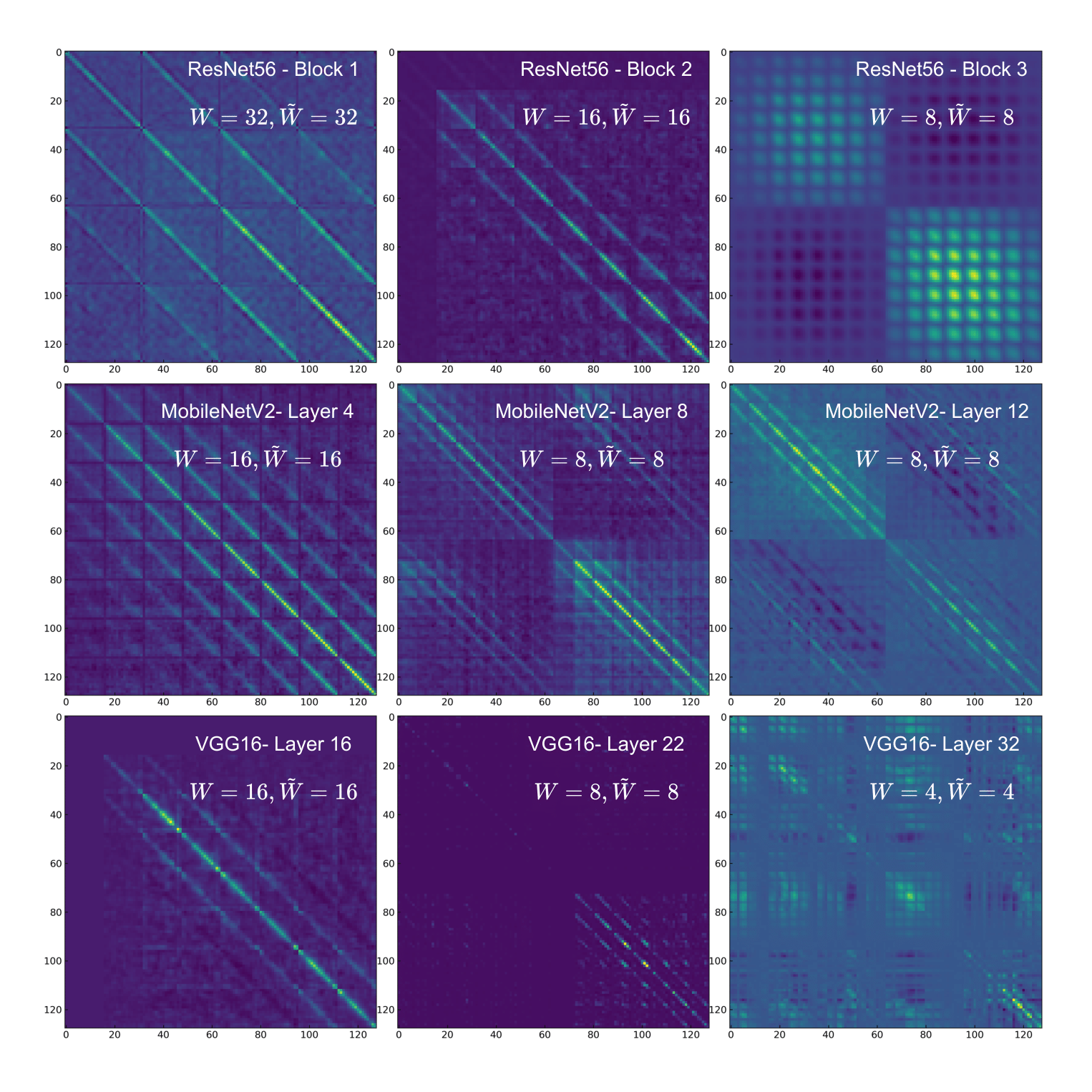}
    \caption{\small{Analysis of the covariance matrix for multiple models and split layers using the CIFAR-10 test set (model and split layer indicated in the first line of each subfigure). For each case, both the ground-truth spatial width \( W \) and the estimated width \( \tilde{W} \) are reported. The analysis was performed using a batch size of 512. The estimated shape can also be visually verified by digitally zooming into the axes of each subfigure.}
}
    \label{fig:correlation-patterns}
\end{figure}

The approach presented in Section~\ref{s:shape_analysis} has been tested across various settings. In particular, Figure~\ref{fig:correlation-patterns} shows representations of the covariance matrices (zoomed in the top left corner for better visualization, as done in Figure \ref{fig:correlation}) for different models and split points, from which the intermediate edge features are extracted. 
Each subfigure includes details such as the model name and the depth of the split point. Additionally, we report both the true spatial width \( W \) and the estimated width \( \tilde{W} \), which is inferred using peak analysis on the autocorrelation of the covariance matrix. While the peak detection is performed automatically, visual inspection of the matrix, by digitally zooming into the axis labels, also enables a quick and intuitive verification of the estimated dimension.
Notably, in all the evaluated cases, the estimated value \( \tilde{W} \) matches the ground-truth width \( W \), confirming the high effectiveness of the approach. For this analysis, the covariance matrices were computed using a batch of 512 feature vectors.

\begin{figure}[h!]
\centering
  \begin{subfigure}{0.7\textwidth}
  \centering
    \includegraphics[width=\textwidth]{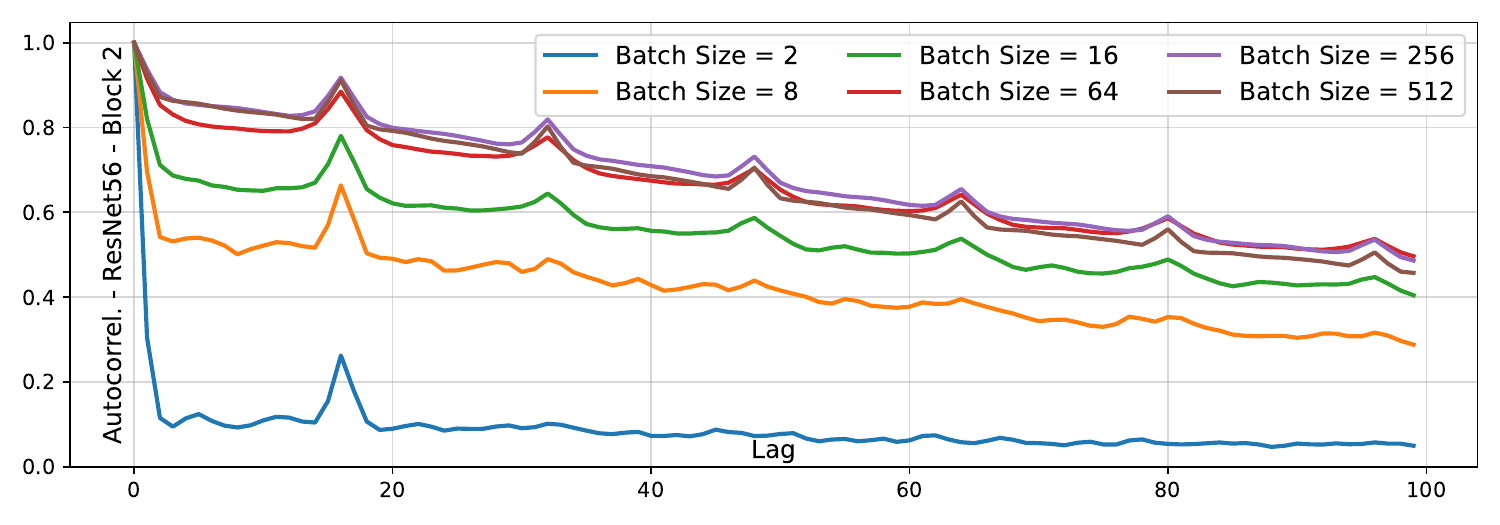}
  \end{subfigure}
  \begin{subfigure}{0.7\textwidth}
  \centering
    \includegraphics[width=\textwidth]{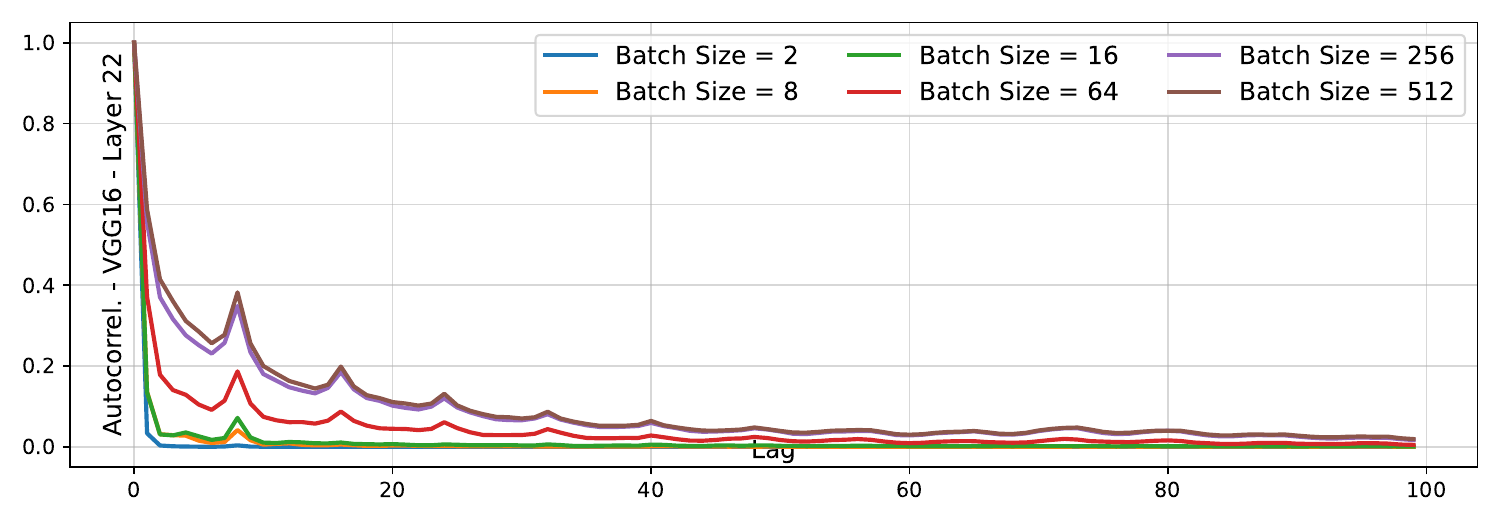}
  \end{subfigure}
  \begin{subfigure}{0.7\textwidth}
  \centering
    \includegraphics[width=\textwidth]{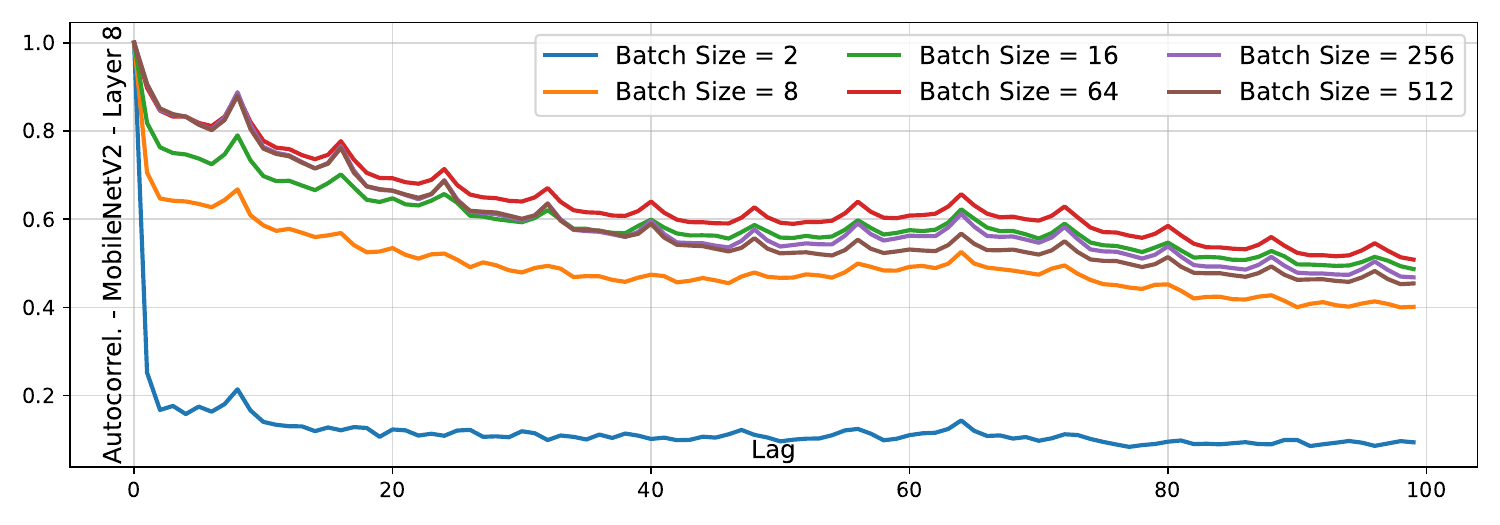}
  \end{subfigure}
  \caption{\small{Analysis of the peak patterns obtained from the autocorrelation curves (normalized in the plots for clearer visual comparison) for different models, with split points selected around the halfway depth of each architecture. In this analysis, we explore the impact of using different batch sizes on the covariance-based shape estimation. The results highlight how smaller batch sizes can negatively affect the clarity of the autocorrelation signal, making the shape estimation less reliable.
}}
\label{fig:edge_estimation_batch}
\end{figure}

To further investigate how the batch size impacts the reliability of the estimation, we repeated the analysis using smaller batches (2, 8, 16, 64, 256, 512). Specifically, we considered intermediate features extracted from different models at split points approximately halfway through their depth, as shown in the y-label of the plots. As illustrated in Figure~\ref{fig:edge_estimation_batch}, reducing the number of samples in the batch can lead to a noisier autocorrelation profile and degraded peak clarity, making the estimation less reliable (e.g., see the blue curve in MobileNetV2, Layer 8, third plot). Conversely, using larger batches provides smoother covariance trends, where the periodic structure of the peaks becomes more evident and supports more accurate shape estimation.

Given the identification of the edge feature shape in our previous experiments (even using a number of samples lower than the total number of queries specific in the experimental settings for doing the distillation part), we now assume this task to be successfully accomplished. In the following experiments, we therefore focus on evaluating the transferability and performance of surrogate models under the assumption that the shape of the intermediate features is known.

\subsection{Evaluation of query-based attacks}
\label{ss:query_based}
These experiments study the advantages of exploiting intermediate features for training surrogates in the context of state-of-the-art query-based black-box attacks, which combine the use of target model queries for gradient estimation with the surrogate model. Specifically, we consider GFCS \cite{lord2022attacking}, Simba-ODS \cite{guo2019simple}, P-RGF \cite{cheng2019improving}, and ODS-RGF \cite{guo2019subspace, cheng2019improving}, which have been set up and implemented according to the specifications in \cite{lord2022attacking}, and provide various analyses based on the $\ell_2$-bounded attacks. In particular, to evaluate the performance of using the proposed surrogate model with features distillation with these attacks we examine the (i) attack success rate; (ii) the number of target queries required to accomplish the attack; and (iii) maximum attack magnitude ($\epsilon$) in unbounded settings for the last experiments (where the method continues iterative steps until achieving an attack without projection in a $\epsilon$-ball).

\begin{figure*}[h]
\centering
  \begin{subfigure}{0.9\textwidth}
  \centering
    \includegraphics[width=1\linewidth]{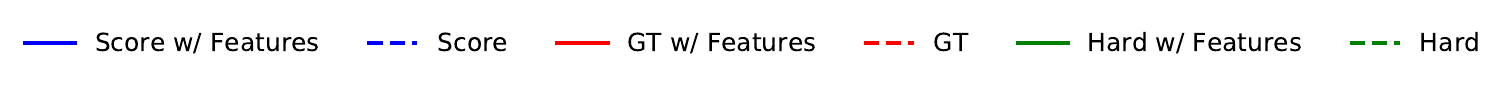}
  \end{subfigure}%
  \hfill
  \begin{subfigure}{\textwidth}
  \centering
    \includegraphics[width=1\linewidth]{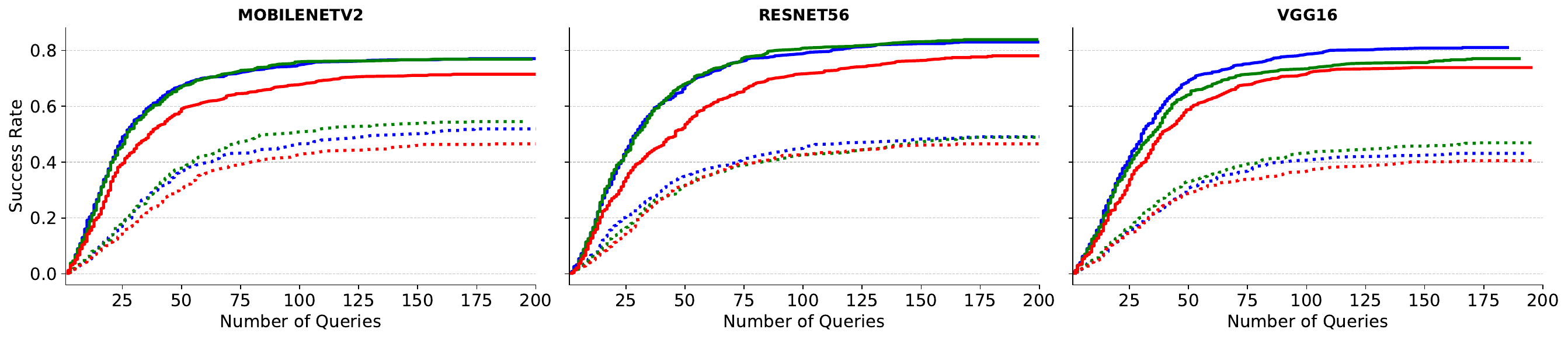}
    \caption{}
  \end{subfigure}
  \begin{subfigure}{\textwidth}
  \centering
    \includegraphics[width=1\linewidth]{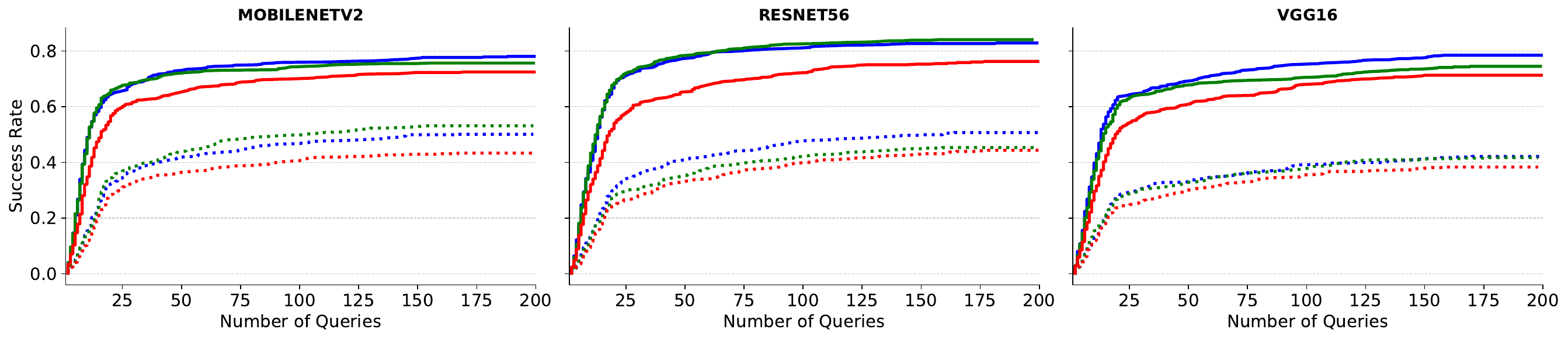}
    \caption{}
  \end{subfigure}
  \caption{\small{Attack success rates as a function of the number of queries for the GFCS attack (a) and Simba-ODS attack (b). For both attacks, we evaluated six different configurations of threat models to highlight the benefits of incorporating intermediate features in the training of the surrogate model.}}
  \label{fig:simba_gfcs_eval}
\end{figure*}

\paragraph{Attack success eate with the number of queries} Figure \ref{fig:simba_gfcs_eval} reports the attack success rate as a function of the number of queries used for the GFCS (a) and (c) and Simba-ODS Attacks (b) and (d), when considering the ResNet56 and the MobileNetV2 as target models, respectively. The surrogate model instead, as specified in the experimental settings, is VGG16. The tests consider three different settings to, highlight the contribution of the distillation depending on the knowledge of the target's output score distinguishing between: full score matching (score), hard prediction matching (hard), and no output availability (label). For these tests, we consider a maximum \(\epsilon\) of 1.50, while \(\alpha\) is set to 0.1. The same settings apply to both Simba-ODS and GFCS. 

As observed, the use of the proposed approach with intermediate features distillation yields significant benefits, enabling even the scenario without known output predictions to outperform, in almost all cases, with a 20\% higher attack success rate at more than 50 queries, compared to a surrogate model trained to mimic only the output score. Furthermore, even with a limited budget of queries (e.g., 25), we observe a significant improvement in attack effectiveness, as indicated by the initial parts of the curves.


\paragraph{Analysis under different \(\epsilon\) values} 
We also examined the impact of varying the perturbation constraint \(\epsilon\) on the transferability of surrogate models. 
To this end, Table~\ref{tab:epsilon_eval} reports the Success Rate (SR) of the attack for GFCS (top) and Simba-ODS (bottom), along with the average number of queries (denoted as "Queries" in the table for simplicity), computed only over the samples for which the attack was successful. 
If the attack does not succeed within a maximum of 100 queries, it is considered a failure.

As expected, smaller \(\epsilon\) values generally make black-box attacks more challenging, often requiring a larger number of queries. 
However, surrogate models trained with intermediate feature exploitation consistently achieve better results with fewer queries, as clearly shown across all three versions.

In particular, at a small \(\epsilon = 0.25\), all tested surrogate models require a high number of queries, reflecting the inherent difficulty of performing black-box attacks under tight perturbation constraints. Conversely, at \(\epsilon = 0.5\), the surrogate model trained with feature distillation demonstrates significantly improved transferability, resulting in a notable reduction in the number of queries required, unlike surrogate models that do not incorporate intermediate feature knowledge.

A similar trend can be observed for both Simba-ODS and GFCS.


  

\begin{table*}[!h]
\centering
\begin{minipage}{0.3\textwidth}
\centering
\caption*{GT Attacks}
\resizebox{1.0\textwidth}{!}{ %
\begin{tabular}{l|cc|ccc}  
\toprule
$\epsilon$ & \multicolumn{2}{c}{SR $\uparrow$} & \multicolumn{2}{c}{Queries $\downarrow$} \\
& w/ & w/o & w/ & w/o \\
\midrule
\multicolumn{5}{c}{\textbf{MobileNetv2}} \\
0.25 & \textbf{0.554} & 0.207 & 13.25 & \textbf{10.58} \\
0.5  & \textbf{0.793} & 0.565 & \textbf{10.08} & 23.35 \\
1.0  & \textbf{0.870} & 0.739 & \textbf{9.74} & 14.56 \\
1.5  & \textbf{0.848} & 0.793 & \textbf{7.64} & 15.84 \\
\midrule
\multicolumn{5}{c}{\textbf{ResNet56}} \\
0.25 & \textbf{0.446} & 0.239 & \textbf{11.39} & 21.45 \\
0.5  & \textbf{0.826} & 0.402 & \textbf{15.49} & 22.86 \\
1.0  & \textbf{0.957} & 0.674 & \textbf{9.30} & 23.32 \\
1.5  & \textbf{0.989} & 0.772 & \textbf{9.19} & 25.49 \\
\midrule
\multicolumn{5}{c}{\textbf{VGG16}} \\
0.25 & \textbf{0.355} & 0.215 & \textbf{9.79} & 14.55 \\
0.5  & \textbf{0.656} & 0.387 & \textbf{7.15} & 20.61 \\
1.0  & \textbf{0.882} & 0.656 & \textbf{14.94} & 23.31 \\
1.5  & \textbf{0.892} & 0.774 & \textbf{14.46} & 23.79 \\
\bottomrule
\end{tabular}
}
\end{minipage}
\begin{minipage}{0.3\textwidth}
\centering
\caption*{Hard Attacks}
\resizebox{1.0\textwidth}{!}{ %
\begin{tabular}{l|cc|ccc}
\toprule
$\epsilon$ & \multicolumn{2}{c}{SR $\uparrow$} & \multicolumn{2}{c}{Queries $\downarrow$} \\
& w/  & w/o & w/ & w/o \\
\midrule
\multicolumn{5}{c}{\textbf{MobileNetv2}} \\
0.25 & \textbf{0.424} & 0.185 & \textbf{13.15} & 19.41 \\
0.5  & \textbf{0.739} & 0.413 & \textbf{10.47} & 14.95 \\
1.0  & \textbf{0.870} & 0.652 & \textbf{12.51} & 20.97 \\
1.5  & \textbf{0.880} & 0.717 & \textbf{12.58} & 19.79 \\
\midrule
\multicolumn{5}{c}{\textbf{ResNet56}} \\
0.25 & \textbf{0.435} & 0.207 & \textbf{20.70} & 29.26 \\
0.5  & \textbf{0.728} & 0.380 & 22.97 & \textbf{19.00} \\
1.0  & \textbf{0.935} & 0.663 & \textbf{15.10} & 23.95 \\
1.5  & \textbf{0.957} & 0.772 & \textbf{15.69} & 22.04 \\
\midrule
\multicolumn{5}{c}{\textbf{VGG16}} \\
0.25 & \textbf{0.312} & 0.194 & \textbf{6.24} & 19.72 \\
0.5  & \textbf{0.645} & 0.376 & 15.93 & \textbf{20.71} \\
1.0  & \textbf{0.860} & 0.624 & \textbf{15.82} & 25.16 \\
1.5  & \textbf{0.892} & 0.710 & \textbf{18.53} & 26.11 \\
\bottomrule
\end{tabular}
}
\end{minipage}
\begin{minipage}{0.3\textwidth}
\centering
\caption*{Score Attacks}
\resizebox{1.0\textwidth}{!}{ %
\begin{tabular}{l|cc|ccc}
\toprule
$\epsilon$ & \multicolumn{2}{c}{SR $\uparrow$} & \multicolumn{2}{c}{Queries $\downarrow$} \\
& w/ & w/o & w/ & w/o \\
\midrule
\multicolumn{5}{c}{\textbf{MobileNetv2}} \\
0.25 & \textbf{0.576} & 0.185 & 7.94 & \textbf{4.76} \\
0.5  & \textbf{0.793} & 0.522 & \textbf{7.23} & 21.17 \\
1.0  & \textbf{0.891} & 0.707 & \textbf{9.78} & 21.71 \\
1.5  & \textbf{0.880} & 0.783 & \textbf{7.86} & 24.04 \\
\midrule
\multicolumn{5}{c}{\textbf{ResNet56}} \\
0.25 & \textbf{0.446} & 0.250 & 11.24 & \textbf{11.00} \\
0.5  & \textbf{0.848} & 0.435 & \textbf{14.24} & 17.20 \\
1.0  & \textbf{0.967} & 0.696 & \textbf{10.80} & 25.20 \\
1.5  & \textbf{0.978} & 0.772 & \textbf{9.60} & 20.15 \\
\midrule
\multicolumn{5}{c}{\textbf{VGG16}} \\
0.25 & \textbf{0.376} & 0.183 & \textbf{6.09} & 12.59 \\
0.5  & \textbf{0.763} & 0.355 & 16.35 & \textbf{15.64} \\
1.0  & \textbf{0.892} & 0.634 & \textbf{13.20} & 22.03 \\
1.5  & \textbf{0.946} & 0.796 & \textbf{17.80} & 27.97 \\
\bottomrule
\end{tabular}
}
\end{minipage}

\begin{minipage}{0.30\textwidth}
\resizebox{1.0\textwidth}{!}{ %
\centering
\begin{tabular}{l|cc|cc}
\toprule
$\epsilon$ & \multicolumn{2}{c}{SR $\uparrow$} & \multicolumn{2}{c}{Queries $\downarrow$} \\
& w/ & w/o & w/ & w/o \\
\midrule
\multicolumn{5}{c}{\textbf{MobileNetv2}} \\
0.25 & \textbf{0.533} & 0.217 & 16.1 & \textbf{14.8} \\
0.5 & \textbf{0.793} & 0.543 & \textbf{15.78} & 26.22 \\
1.0 & \textbf{0.902} & 0.75 & \textbf{19.61} & 26.51 \\
1.5 & \textbf{0.902} & 0.804 & \textbf{16.92} & 30.65 \\
\midrule
\multicolumn{5}{c}{\textbf{ResNet56}} \\
0.25 & \textbf{0.446} & 0.228 & 21.0 & \textbf{18.19} \\
0.5 & \textbf{0.804} & 0.402 & \textbf{20.97} & 22.41 \\
1.0 & \textbf{0.957} & 0.674 & \textbf{17.84} & 32.9 \\
1.5 & \textbf{0.989} & 0.793 & \textbf{22.25} & 33.78 \\
\midrule
\multicolumn{5}{c}{\textbf{VGG16}} \\
0.25 & \textbf{0.323} & 0.204 & \textbf{8.87} & 16.95 \\
0.5 & \textbf{0.677} & 0.376 & \textbf{14.19} & 18.09 \\
1.0 & \textbf{0.946} & 0.677 & \textbf{21.38} & 32.65 \\
1.5 & \textbf{0.957} & 0.785 & \textbf{22.7} & 30.67 \\
\bottomrule
\end{tabular}
}
\end{minipage}
\begin{minipage}{0.30\textwidth}
\centering
\resizebox{1.0\textwidth}{!}{ %
\begin{tabular}{l|cc|cc}
\toprule
$\epsilon$ & \multicolumn{2}{c}{SR $\uparrow$} & \multicolumn{2}{c}{Queries $\downarrow$} \\
& w/ & w/o & w/ & w/o \\
\midrule
\multicolumn{5}{c}{\textbf{MobileNetv2}} \\
0.25 & \textbf{0.402} & 0.174 & \textbf{13.54} & 18.19 \\
0.5 & \textbf{0.772} & 0.413 & \textbf{16.2} & 23.29 \\
1.0 & \textbf{0.859} & 0.652 & \textbf{17.2} & 30.7 \\
1.5 & \textbf{0.88} & 0.739 & \textbf{19.05} & 34.03 \\
\midrule
\multicolumn{5}{c}{\textbf{ResNet56}} \\
0.25 & \textbf{0.413} & 0.185 & \textbf{16.13} & 27.06 \\
0.5 & \textbf{0.696} & 0.38 & 22.77 & \textbf{22.37} \\
1.0 & \textbf{0.946} & 0.707 & \textbf{28.2} & 32.74 \\
1.5 & \textbf{0.935} & 0.793 & \textbf{25.02} & 35.53 \\
\midrule
\multicolumn{5}{c}{\textbf{VGG16}} \\
0.25 & \textbf{0.323} & 0.183 & \textbf{10.57} & 19.35 \\
0.5 & \textbf{0.667} & 0.376 & 20.69 & \textbf{20.49} \\
1.0 & \textbf{0.882} & 0.613 & \textbf{21.37} & 25.81 \\
1.5 & \textbf{0.892} & 0.763 & \textbf{20.19} & 37.38 \\
\bottomrule
\end{tabular}
}
\end{minipage}
\begin{minipage}{0.30\textwidth}
\resizebox{1.0\textwidth}{!}{ %
\centering
\begin{tabular}{l|cc|cc}
\toprule
$\epsilon$ & \multicolumn{2}{c}{SR $\uparrow$} & \multicolumn{2}{c}{Queries $\downarrow$} \\
& w/ & w/o & w/ & w/o \\
\midrule
\multicolumn{5}{c}{\textbf{MobileNetv2}} \\
0.25 & \textbf{0.522} & 0.217 & \textbf{15.42} & 17.25 \\
0.5 & \textbf{0.772} & 0.5 & \textbf{12.8} & 25.09 \\
1.0 & \textbf{0.902} & 0.728 & \textbf{16.64} & 30.97 \\
1.5 & \textbf{0.902} & 0.772 & \textbf{16.08} & 32.07 \\
\midrule
\multicolumn{5}{c}{\textbf{ResNet56}} \\
0.25 & \textbf{0.435} & 0.25 & \textbf{10.68} & 13.83 \\
0.5 & \textbf{0.804} & 0.435 & \textbf{17.34} & 19.45 \\
1.0 & \textbf{0.957} & 0.717 & \textbf{17.06} & 33.36 \\
1.5 & \textbf{0.989} & 0.848 & \textbf{19.67} & 39.32 \\
\midrule
\multicolumn{5}{c}{\textbf{VGG16}} \\
0.25 & \textbf{0.387} & 0.194 & \textbf{14.53} & 18.22 \\
0.5 & \textbf{0.731} & 0.387 & \textbf{18.19} & 20.56 \\
1.0 & \textbf{0.935} & 0.656 & \textbf{19.51} & 28.51 \\
1.5 & \textbf{0.946} & 0.796 & \textbf{19.07} & 35.45 \\
\bottomrule
\end{tabular}
}
\end{minipage}
\caption{\small{Comparison of black-box attacks (GFCS, top; Simba-ODS, bottom) considering the use of feature distillation (w/) under different output threat models: GT, Hard, and Score attacks, evaluated across models and perturbation budgets ($\epsilon$). Best values in terms of success rate (SR) and average number of queries (Queries) per row are highlighted in bold. Surrogate models trained without feature distillation are denoted as w/o.}}
\label{tab:epsilon_eval}
\end{table*}

\begin{figure*}[h]
    \begin{subfigure}{0.62\textwidth}
    \centering
            \resizebox{1.0\textwidth}{!}{ %
            \begin{tabular}{@{}lccc|ccc||ccc|ccc@{}}
            \toprule
            & \multicolumn{3}{c}{Score} & \multicolumn{3}{c}{\textbf{FD+Score}} & \multicolumn{3}{c}{Hard} & \multicolumn{3}{c}{\textbf{FD+Hard}} \\
            \cmidrule(lr){2-4} \cmidrule(lr){5-7} \cmidrule(lr){8-10} \cmidrule(l){11-13}
            Model & SR$\uparrow$  & AP$\downarrow$  & AQ$\downarrow$ &SR$\uparrow$  & AP$\downarrow$  & AQ$\downarrow$  & SR$\uparrow$ & AP$\downarrow$  & AQ$\downarrow$  & SR$\uparrow$ & AP$\downarrow$  & AQ$\downarrow$ \\
            \midrule
            \multicolumn{13}{c}{\small VGG16} \\
            \midrule
            Simba-ODS & 0.80 & 0.9 & 17 & \textbf{0.99} & \textbf{0.57} & \textbf{6} & 0.76 & 0.95 & 22 & \textbf{0.95} & \textbf{0.59} & \textbf{7} \\
            GFCS & 0.81 & 0.92 & 17 & \textbf{0.98} & \textbf{0.53} & \textbf{6} & 0.74 & 0.93 & 21 & \textbf{0.96} & \textbf{0.58} & \textbf{7} \\
            P-RGF & 0.86 & 0.93 & 17 & \textbf{0.98} & \textbf{0.55} & \textbf{6} & 0.77 & 0.98 & 22 & \textbf{0.94} & \textbf{0.58} & \textbf{7} \\
            RGF-ODS & 0.85 & 0.91 & 17 & \textbf{0.99} & \textbf{0.55} & \textbf{6} & 0.79 & 0.95 & 21 & \textbf{0.94} & \textbf{0.56} & \textbf{7} \\
            \midrule
            \multicolumn{13}{c}{\small ResNet56} \\
            \midrule
            Simba-ODS & 0.86 & 0.8 & 16.5 & \textbf{0.99} & \textbf{0.43} & \textbf{6.5} & 0.86 & 0.76 & 12 & \textbf{0.99} & \textbf{0.42} & \textbf{6} \\
            GFCS & 0.86 & 0.78 & 17.5 & \textbf{0.98} & \textbf{0.42} & \textbf{6.5} & 0.86 & 0.77 & 12 & \textbf{0.99} & \textbf{0.42} & \textbf{6} \\
            P-RGF & 0.86 & 0.78 & 16.5 & \textbf{0.99} & \textbf{0.43} & \textbf{6.5} & 0.85 & 0.76 & 12 & \textbf{0.98} & \textbf{0.42} & \textbf{6} \\
            RGF-ODS & 0.85 & 0.79 & 16 & \textbf{0.98} & \textbf{0.43} & \textbf{6.5} & 0.85 & 0.74 & 12 & \textbf{0.97} & \textbf{0.42} & \textbf{6} \\
            \midrule
            \multicolumn{13}{c}{\small MobileNetV2} \\
            \midrule
            Simba-ODS & 0.86 & 0.75 & 11 & \textbf{1.0} & \textbf{0.38} & \textbf{5} & 0.9 & 0.74 & 11 & \textbf{1.0} & \textbf{0.4} & \textbf{6} \\
            GFCS & 0.89 & 0.74 & 14 & \textbf{0.99} & \textbf{0.38} & \textbf{5} & 0.92 & 0.77 & 11 & \textbf{1.} & \textbf{0.4} & \textbf{6} \\
            P-RGF & 0.86 & 0.77 & 12 & \textbf{0.99} & \textbf{0.39} & \textbf{5} & 0.91 & 0.76 & 11 & \textbf{0.98} & \textbf{0.39} & \textbf{6} \\
            RGF-ODS & 0.86 & 0.77 & 13 & \textbf{0.99} & \textbf{0.39} & \textbf{5} & 0.87 & 0.7 & 11 & \textbf{0.99} & \textbf{0.4} & \textbf{6} \\
            \bottomrule \\
            \end{tabular} }
            \caption{}
            \label{tab:attacks_evaluation}
    \end{subfigure}
    \begin{subfigure}{0.35\textwidth}
        \centering
        \begin{subfigure}{0.9\textwidth}
        \includegraphics[width=\textwidth]{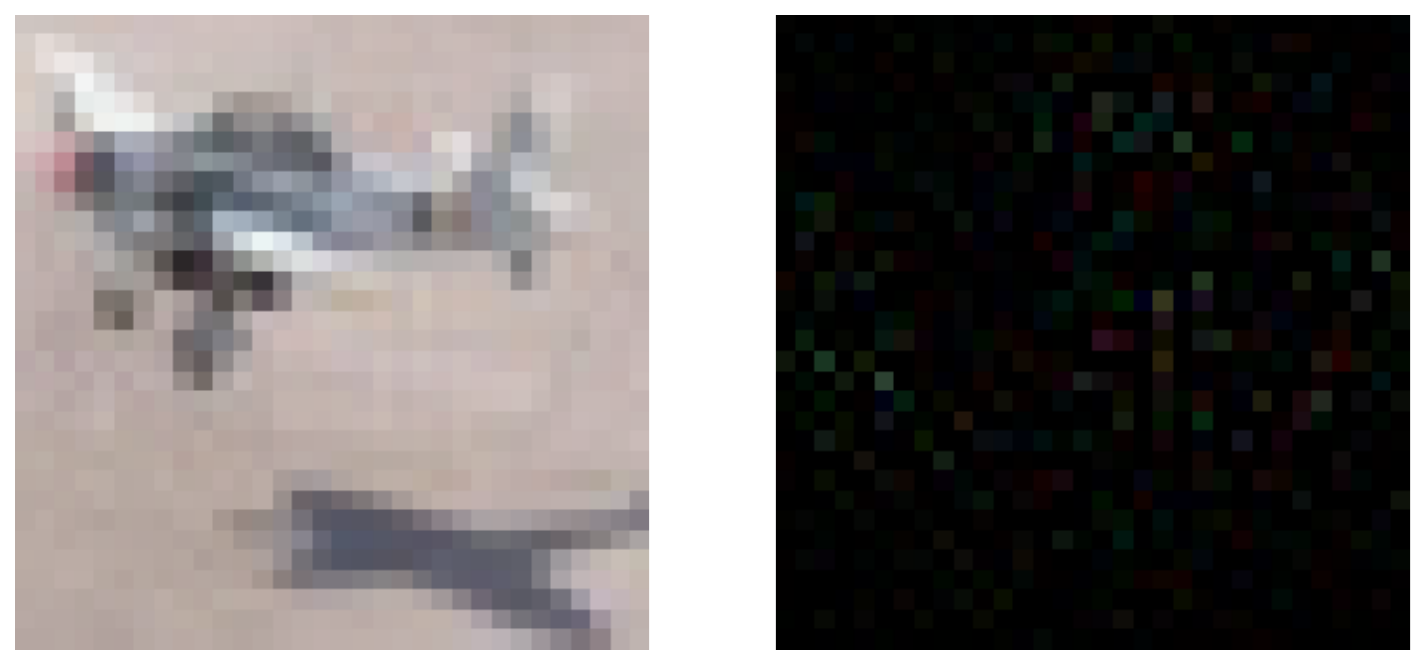} 
        \caption*{Surrogate w/ FD+Score}
        \end{subfigure}
        \quad\quad
        \begin{subfigure}{0.9\textwidth}
        \includegraphics[width=\textwidth]{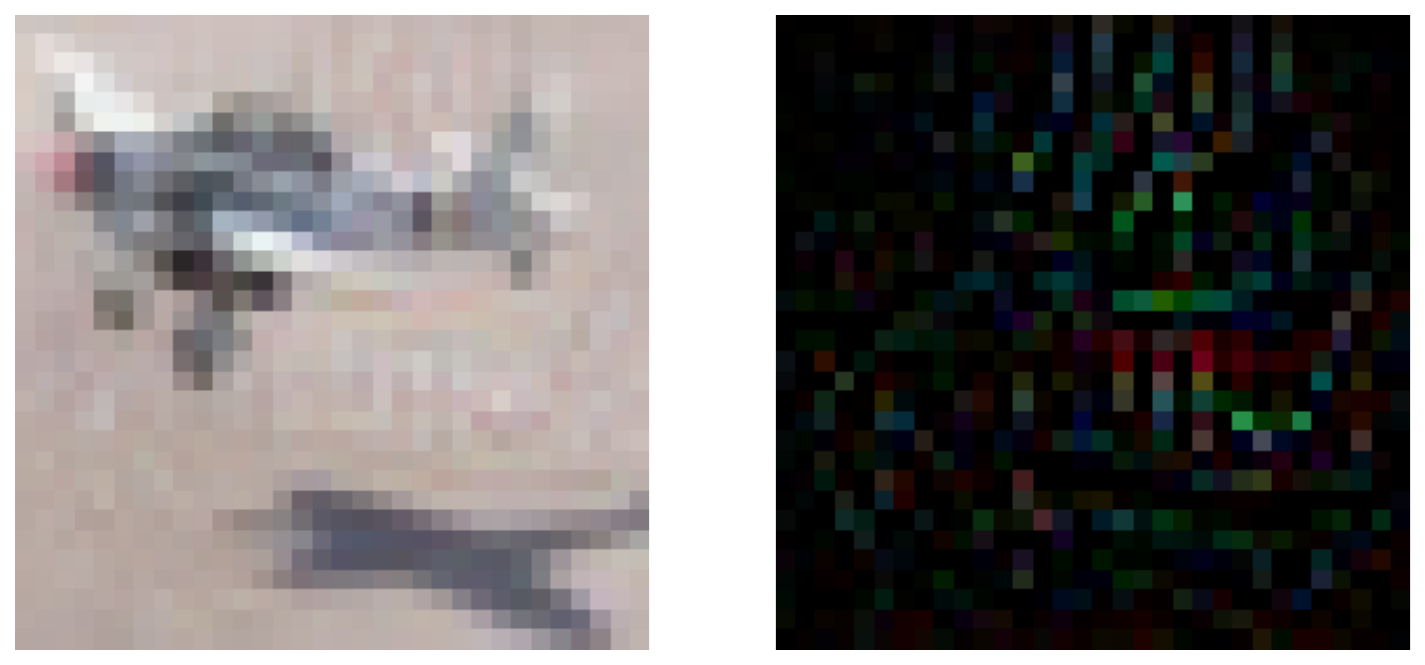} 
        \caption*{Surrogate w/ Score}
        \end{subfigure}
        \caption{}
    \end{subfigure}
    \caption{\small{(a) Results of different black-box attacks under unbounded constraints ($\epsilon = \infty$) including Simba-ODS, GFCS, P-RGF, and RGF-ODS, using surrogate models trained under various threat scenarios, with and without feature distillation (FD). The reported metrics include the Success Rate (SR), Average Queries (AQ), and Average Perturbation magnitude (AP), computed using the $\ell_{2}$ norm. (b) Visual comparison of adversarial examples and their corresponding perturbations (rescaled $\times 10$) generated using GFCS under unbounded settings ($\epsilon = \infty$).}}
    \label{fig:airplane}
\end{figure*}

\paragraph{Analysis of perturbation in unbounded settings} Table~\ref{tab:attacks_evaluation} presents an evaluation of attack strategies in an unbounded setting, where the perturbation constraint \( \epsilon \) is set to infinity. This implies that the attack does not restrict the perturbation magnitude with a predefined upper bound, beyond the standard clipping constraint imposed by the image format. 

For this analysis, we fix the maximum number of queries to 25 and report the following metrics: attack success rate, average perturbation magnitude (in terms of \( \ell_2 \) norm), and average number of queries used. The latter two statistics are computed only over the subset of samples where the attack succeeds. The surrogate models used for the attacks are trained under different configurations, including with and without feature distillation.
The goal of this analysis is to understand how, in the absence of a constraint on \( \epsilon \), the resulting perturbations may become more perceptible to the human eye. We specifically investigate whether feature distillation affects the visibility of perturbations.

The results show that surrogate models trained with intermediate feature exploitation consistently yield higher attack success rates while also producing lower average perturbation magnitudes. This behavior is observed both in the case where full output scores are available and when only hard labels are used (corresponding to the left and right sections of the table, respectively).

To further illustrate these benefits, Figure~\ref{fig:airplane} shows an example input attacked in an unconstrained setting, using configurations with and without feature distillation (left and right, respectively), with ResNet as the target model. As observed, the injected perturbation is significantly more visible when feature distillation is not used. For better visualization, the perturbation in the right panel is rescaled by a factor of 10 to enhance its visibility to the human eye.




\label{ss:white_box}
\subsection{Impact in white-box surrogate attacks}

We also explored the white-box transferability of surrogate models and the benefits of incorporating feature-level knowledge. Specifically, the Projected Gradient Descent (PGD) attack~\cite{pgd_attack} was applied to the surrogate model, using both $\ell_2$ and $\ell_\infty$ norms, setting $k$, and the number of iterations equal to 20. The resulting adversarial examples, generated using only gradient information from the surrogate, were then evaluated on each target model to assess attack transferability. In the graphs, darker colors represent the transferability (i.e., attack success rate) of surrogate models without feature distillation, while lighter colors indicate the version with feature distillation, highlighting the improvement in success rate.

\begin{figure*}[ht]
  \begin{subfigure}{\textwidth}
  \centering
    \includegraphics[width=0.92\textwidth]{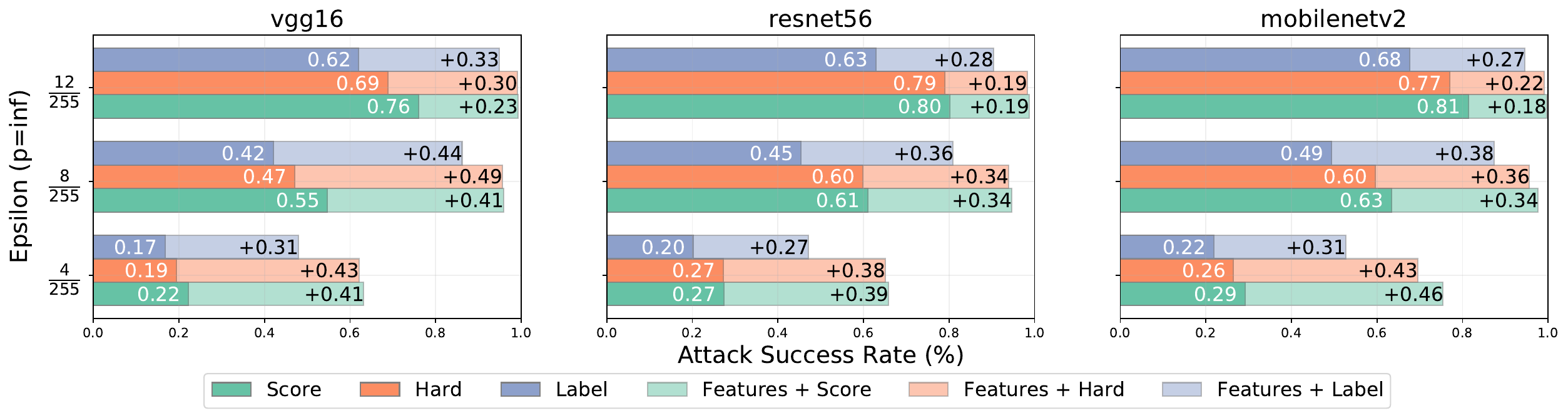}
  \end{subfigure}
  \begin{subfigure}{\textwidth}
  \centering
    \includegraphics[width=0.92\textwidth]{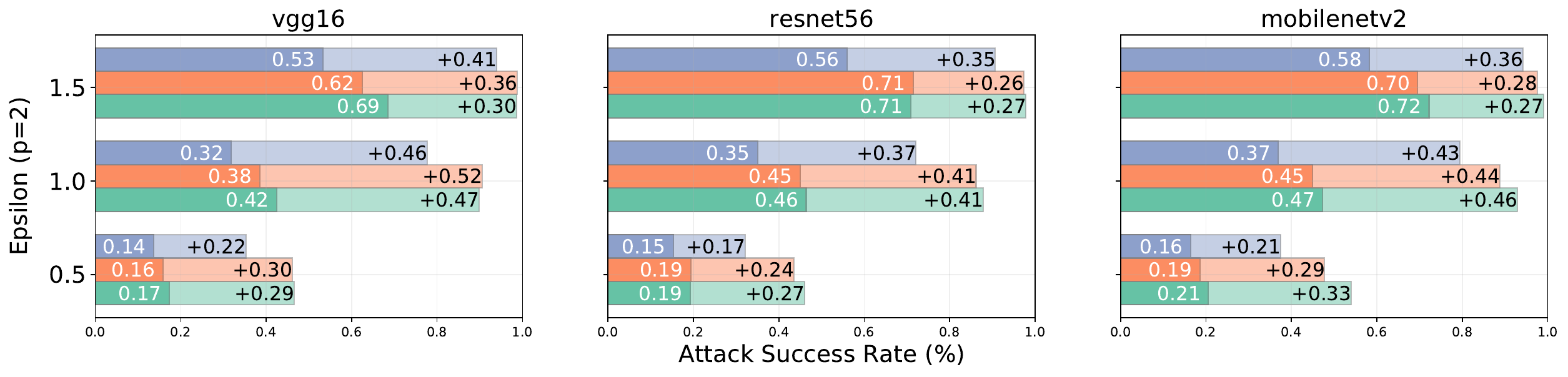}
  \end{subfigure}
  \caption{\small{Impact of White-Box Surrogate Attacks on Pretrained Target Networks Using a VGG-16 Model with Feature Matching at Block-2, Assuming an Edge at Block-1. The first row illustrates the outcomes of PGD \( \ell_{\infty} \) attacks, while the second row reports results for PGD \( \ell_{2} \) norm attacks across different \( \epsilon \) values. These tests consider a threat model with score-level knowledge and investigate the effect of intermediate feature distillation. 
Lighter colors represent attacks with feature distillation, while darker colors correspond to attacks without it.}}
  \label{fig:untarget_white_box}
\end{figure*}

As shown in Figure~\ref{fig:untarget_white_box}, the integration of intermediate feature into the training of the surrogate significantly improves attack success rates under both $\ell_2$  and $\ell_\infty$ norms. Notably, under the \( \ell_{\infty} \) setting, we observe an improvement of 27\% in the worst-case scenario (without access to the target model’s outputs) for a perturbation magnitude of \( \frac{8}{255} \). In the best-case scenario, a success rate increase of up to 51\% is achieved when attacking a ResNet56 model. This substantial improvement is likely due to architectural similarities between the surrogate and the target model.


These results highlight the significant gains in transferability enabled by feature distillation, even in surrogate-only white-box attacks (i.e., without requiring any queries to the target). This underscores the importance of developing robust defense mechanisms, particularly for systems that employ partitioned inference paradigms, which are especially vulnerable to such sophisticated black-box threat models.

%

\subsection{Feature distillation across different layer depths}
\label{ss:split_depth}

While all previous experiments were conducted with fixed split points in both the target and surrogate models, we now investigate how the transferability of surrogate models trained with feature distillation varies with respect to different split points. Specifically, we explore the effect of changing the depth at which the surrogate is split, as well as the impact of the (unknown) split point in the target model. This analysis highlights the importance of aligning the partitioning depth of the surrogate’s encoder (edge part) and decoder (cloud part) with that of the target model, particularly relevant in realistic black-box scenarios where the architecture of the target model is unknown.

\begin{figure}[h]
\centering
  \begin{subfigure}{0.3\textwidth}
  \centering
    \includegraphics[width=1\textwidth]{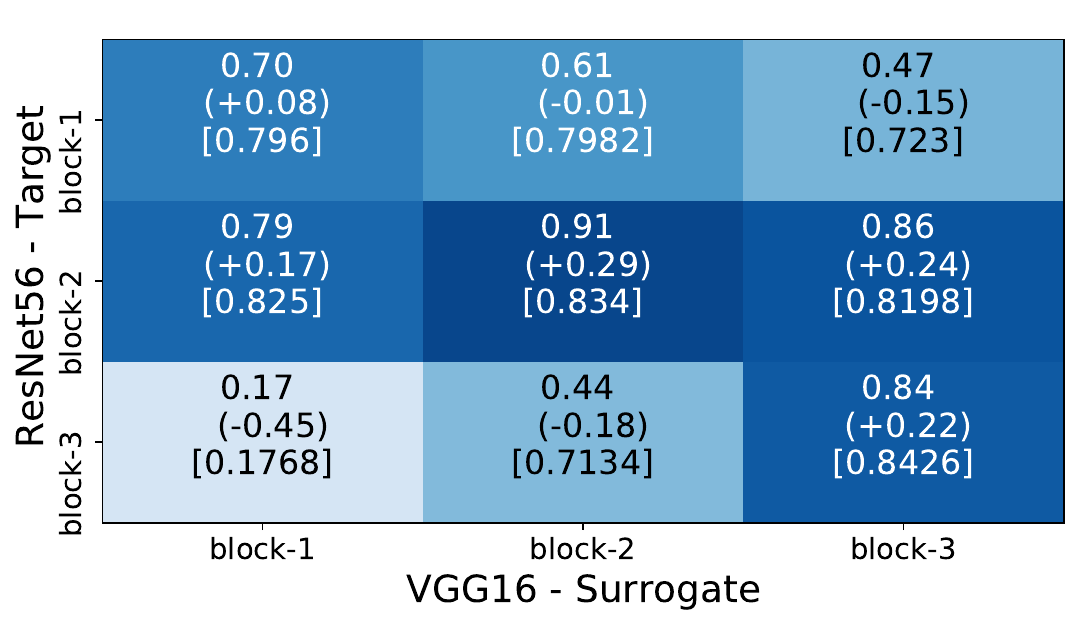}
  \end{subfigure}
  \begin{subfigure}{0.3\textwidth}
  \centering
    \includegraphics[width=1\textwidth]{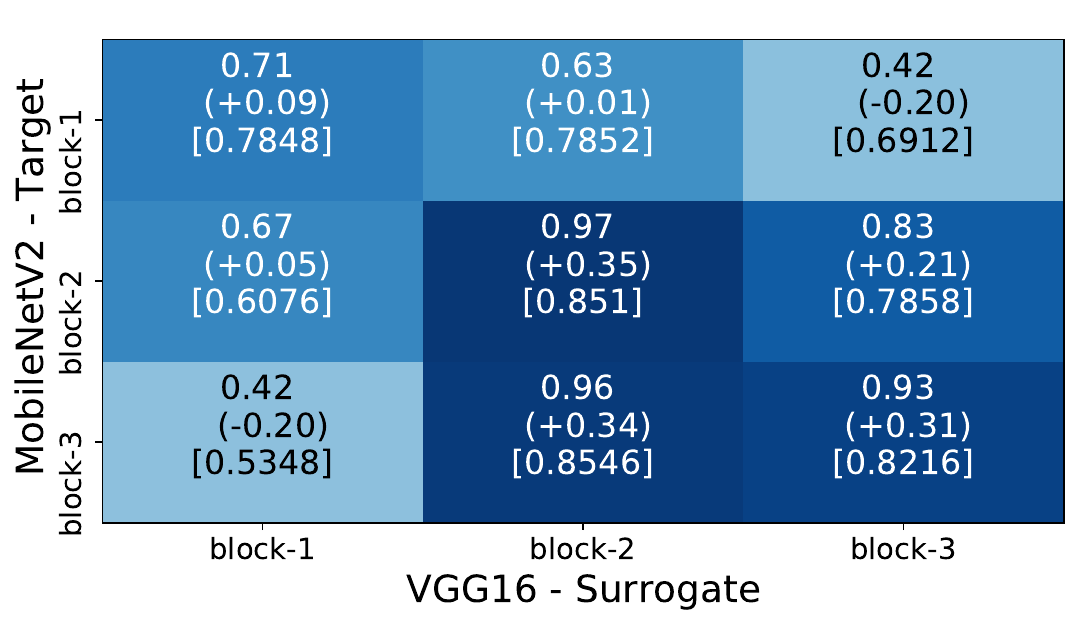}
  \end{subfigure}
   \begin{subfigure}{0.3\textwidth}
  \centering
    \includegraphics[width=1\textwidth]{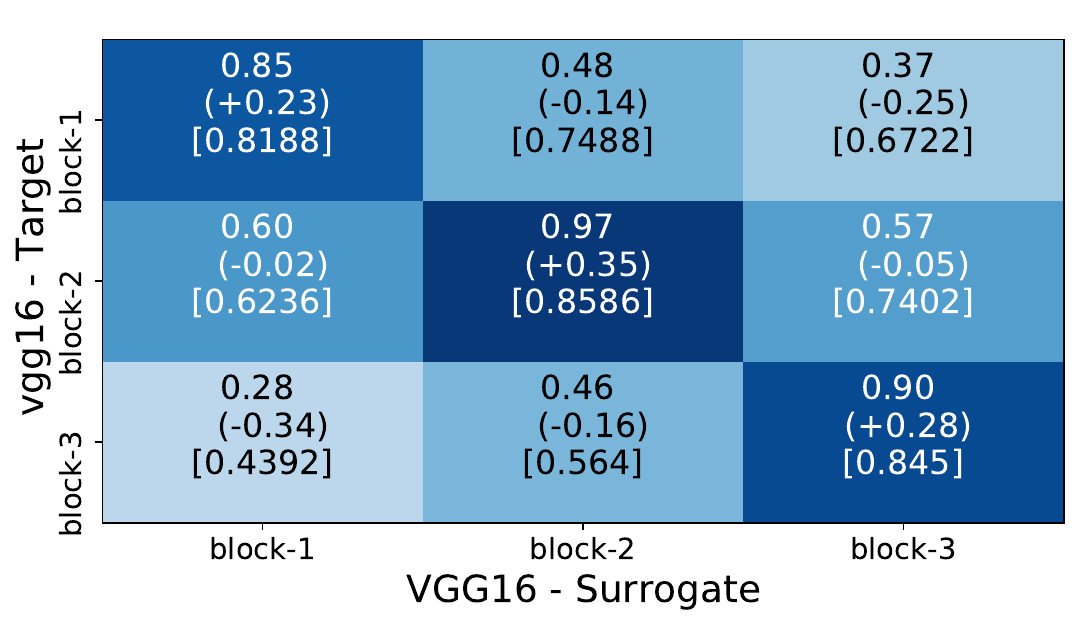}
  \end{subfigure}
  \begin{subfigure}{0.3\textwidth}
  \centering
    \includegraphics[width=1\textwidth]{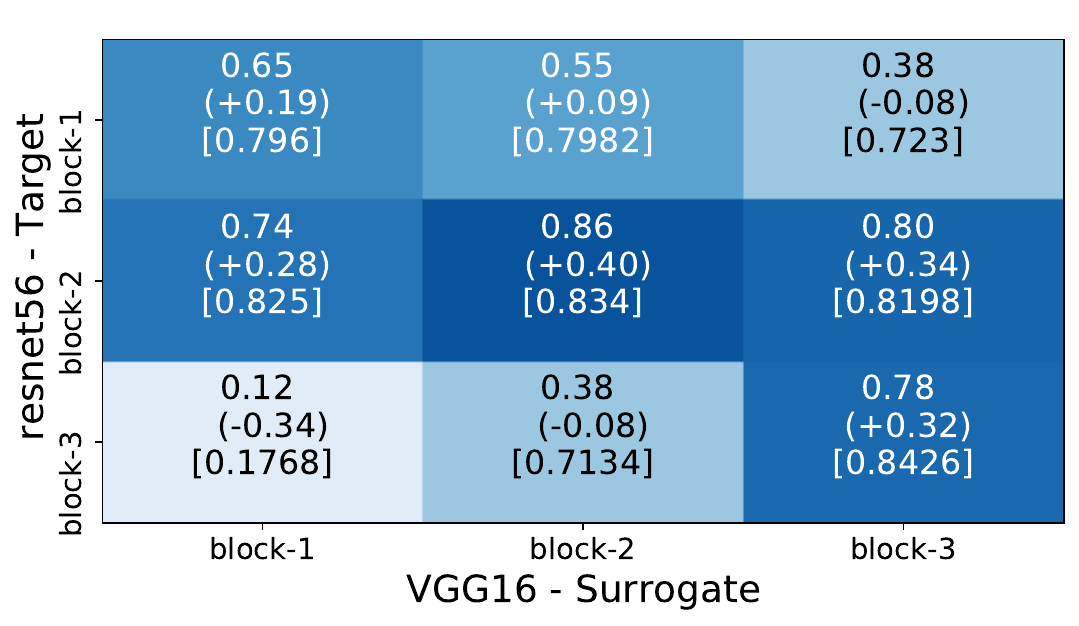}
  \end{subfigure}
  \begin{subfigure}{0.3\textwidth}
  \centering
    \includegraphics[width=1\textwidth]{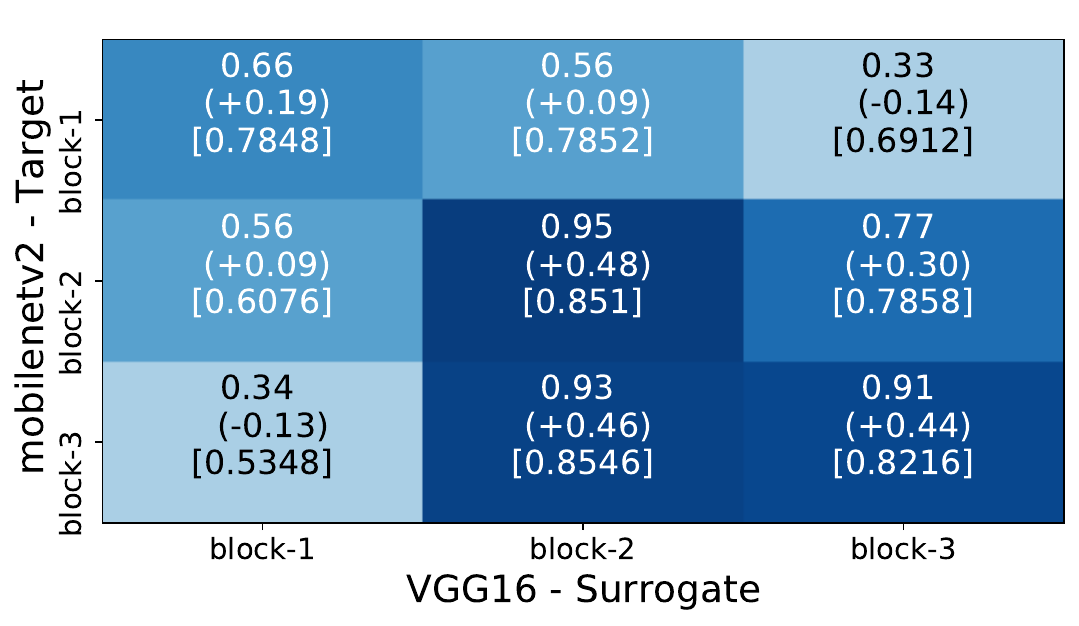}
  \end{subfigure}
   \begin{subfigure}{0.3\textwidth}
  \centering
    \includegraphics[width=1\textwidth]{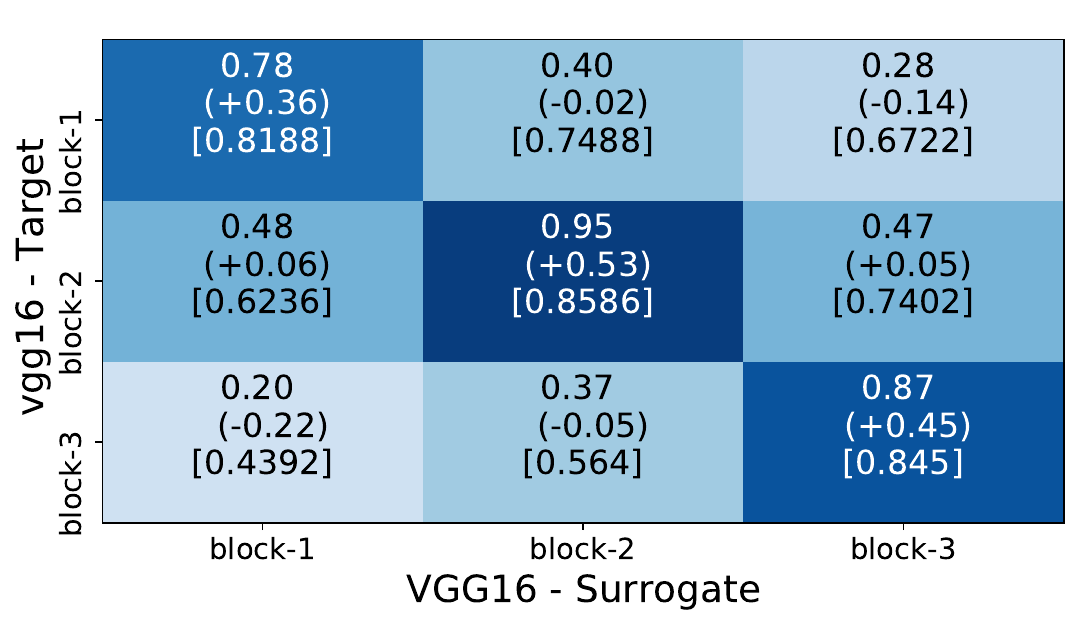}
  \end{subfigure}
  \caption{\small{Analysis of the Effectiveness of Feature Distillation Across Different Layers Using VGG-16 as the Surrogate Model. 
Values inside each box show the attack success rate (\%) for two settings: PGD \(\ell_{\infty}\) with \(\epsilon = \tfrac{8}{255}\) (top) and PGD \(\ell_{\infty}\) with \(\epsilon = 1\) (bottom). Both attacks use \(k = 20\) iterations, employ the feature-distillation strategy, and assume score-only access.
The difference in success rate compared to attacks without feature distillation is shown in parentheses in the middle. 
Square brackets indicate the accuracy on clean test samples.}}
  \label{fig:split_depth}
\end{figure}

To this end, Figure~\ref{fig:split_depth} reports the results obtained using VGG-16 as the reference surrogate model with the proposed auxiliary feature-distillation block. We evaluate the effectiveness of PGD-\(\ell_{\infty}\) attacks with \( \epsilon = \frac{8}{255} \). In each cell of the matrix, the top entry shows the attack success rate for a given combination of split points, along with the difference (in parentheses) with respect to a baseline surrogate model trained without feature distillation. This provides a direct comparison between distillation-aware and standard surrogates. In the bottom entry of each cell, we also report the clean accuracy of the surrogate (with feature distillation) on the test set, which helps contextualize how effective the surrogate is as a classifier and indirectly supports an empirical understanding of how split point selection affects attack transferability.

Notably, we observe that when the surrogate and target models share similar partitioning depths, the attack success rate tends to be higher. However, effective transferability is still maintained even in cases with moderate discrepancies in split depth. This suggests that while alignment between the surrogate and target depth improves performance, the approach is somewhat robust to variations, provided the difference is not too large.
Conversely, when the depth mismatch becomes too much, however, performance degrades. In these cases, misalignment in the feature distillation process can lead to adverse effects on transferability. Specifically, in some configurations (e.g., when the target is split at a deep layer and the surrogate at a very shallow one), the attack success rate drops below that of the non-features-distilled baseline. These cases are clearly shown for the cells in lower-left corner of all the produced matrices in Figure~\ref{fig:split_depth}, where negative differences are reported in parentheses.

\begin{figure}[ht]
\centering
    \includegraphics[width=0.75\columnwidth]{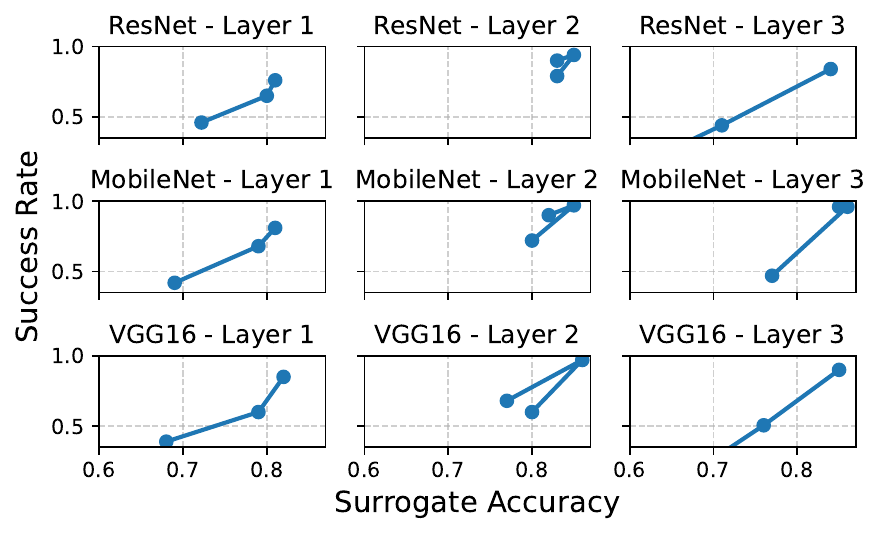}
\caption{\small{Study of the relationship between clean accuracy on the test set and the success rate transferability of the attacks from Figure \ref{fig:split_depth}, considering different splitting points for each configuration on the unknown target side.}}
\label{fig:stat_depth}
\end{figure}

To further emphasize the importance of this analysis and to derive an empirical guideline for selecting the surrogate split depth, particularly in scenarios where the split point of the target model is unknown, we extract a correlation from the previous experiments, as shown in Figure~\ref{fig:stat_depth}. This analysis reveals a clear relationship between the transferability of the surrogate model and its clean accuracy on the test set, since for each target split configuration, the surrogate version considering a split points that get towards a higher results on clean data also bring much more transferability. 
This insight suggests that an attacker can estimate which surrogate configuration is likely to yield higher transferability by simply observing the clean accuracy of different surrogate variants. In practice, the attacker can train multiple surrogate models with different split points and evaluate their classification performance on clean samples (i.e., studying each row of the matrix from the previous analysis, assuming the target split is unknown and fixed). Note that all intermediate features can be collected from the same intercepted batch, so no additional queries are required to perform this analysis.

This provides an empirical approach for selecting the most transferable surrogate configuration from the attacker’s perspective, specifically, identifying the optimal split depth at which to apply feature distillation.
The underlying intuition is as follows: intermediate features extracted from the target's edge part tend to be more semantically rich and discriminative when obtained from deeper layers, while features from shallower layers may retain more low-level details. Consequently, a surrogate model capable of effectively mimicking these features must also possess sufficient depth in both its edge and cloud components. For example, if the target’s edge part is deep, the surrogate's edge part must also be deep enough to extract similarly fine-grained representations. At the same time, a sufficiently deep cloud part is required to process these features and achieve high predictive accuracy.
In this context, the clean accuracy of a surrogate model serves as a useful proxy for the degree of alignment between its internal representations and those of the target. Therefore, selecting the surrogate split point that yields the highest clean accuracy can empirically guide the attacker toward a configuration with stronger attack transferability in black-box settings.

\section{Conclusions and future directions}
\label{s:conclusion}
This work explores the landscape of transferable attacks based on surrogate models in a novel and increasingly relevant scenario: distributed AI paradigms and AI-IoT systems, such as split collaborative learning. In this setting, the attacker is assumed to have access to the intermediate features transmitted between the edge and cloud components of a model. By intercepting this data, the attacker can significantly improve the effectiveness of surrogate models used for transferable evasion attacks.
To enable this, we address key challenges that arise in scenarios where communication between the edge and cloud is exposed, assuming that encryption and network security mechanisms can be bypassed. These challenges include: (i) reconstructing the original feature shape from the vectorized transmitted data, and (ii) adapting classical architectures to enable feature-level distillation for surrogate model training.
Both challenges have been tackled through a statistical analysis based on the covariance matrix of the intercepted features and the design of a lightweight adaptation module that aligns the extracted features with surrogate model layers.

Our experimental results demonstrate that this novel, yet practical, threat model significantly impacts the transferability of surrogate models in both black-box and white-box attack settings. The use of intermediate features proves to be a key factor in enhancing transferability, thereby highlighting a critical vulnerability in distributed AI systems. These findings open new directions for future work in transferability research, particularly regarding feature-level distillation and its role in the attack surface of AI-IoT scenarios.

Future studies could focus on developing defense mechanisms tailored to split inference settings that hinder the attacker’s ability to reconstruct feature shapes, thereby limiting the feasibility of such attacks. This would open promising avenues for studying more robust feature obfuscation strategies and structured defenses that preserve performance while protecting internal representations. On the offensive side, further research could also explore how to design more advanced feature-based attacks, potentially improving attack efficiency or bypassing newly proposed defenses through adaptive surrogate training techniques.

We believe this work provides important insights into the feasibility and effectiveness of exploiting feature exposure in a black-box scenario, potentially constituting other viable threat model in collaborative learning fields. Future research could focus on enhancing the effectiveness of the attacks by exploring more refined feature distillation techniques, as well as variations of the proposed threat model and the attacker's objectives.

{\small
\section*{Acknowledgements}
This work was partially supported by project SERICS (PE00000014) under the MUR (Ministero dell'Università e della Ricerca) National Recovery and Resilience Plan funded by the European Union - NextGenerationEU.}

\bibliographystyle{unsrt}
\bibliography{main}

\begin{thebibliography}{10}

\bibitem{verbraeken2020survey}
Joost Verbraeken, Matthijs Wolting, Jonathan Katzy, Jeroen Kloppenburg, Tim Verbelen, and Jan~S Rellermeyer.
\newblock A survey on distributed machine learning.
\newblock {\em Acm computing surveys (csur)}, 53(2):1--33, 2020.

\bibitem{Kang_collaborative}
Yiping Kang, Johann Hauswald, Cao Gao, Austin Rovinski, Trevor Mudge, Jason Mars, and Lingjia Tang.
\newblock Neurosurgeon: Collaborative intelligence between the cloud and mobile edge.
\newblock {\em SIGARCH Comput. Archit. News}, 45(1):615–629, apr 2017.

\bibitem{distributed_dnn_Teerapittayanon}
Surat Teerapittayanon, Bradley McDanel, and H.T. Kung.
\newblock Distributed deep neural networks over the cloud, the edge and end devices.
\newblock In {\em 2017 IEEE 37th International Conference on Distributed Computing Systems (ICDCS)}, pages 328--339, 2017.

\bibitem{Ko_partitioning}
Jong~Hwan Ko, Taesik Na, Mohammad~Faisal Amir, and Saibal Mukhopadhyay.
\newblock Edge-host partitioning of deep neural networks with feature space encoding for resource-constrained internet-of-things platforms.
\newblock In {\em 2018 15th IEEE International Conference on Advanced Video and Signal Based Surveillance (AVSS)}, pages 1--6, 2018.

\bibitem{Anjos_survey_collaborative}
Julio C. S.~Dos Anjos, Kassiano~J. Matteussi, Fernanda~C. Orlandi, Jorge L.~V. Barbosa, Jorge~S\'{a} Silva, Luiz~F. Bittencourt, and Cl\'{a}udio F.~R. Geyer.
\newblock A survey on collaborative learning for intelligent autonomous systems.
\newblock {\em ACM Comput. Surv.}, 56(4), nov 2023.

\bibitem{biggio2013evasion}
Battista Biggio, Igino Corona, Davide Maiorca, Blaine Nelson, Nedim {\v{S}}rndi{\'c}, Pavel Laskov, Giorgio Giacinto, and Fabio Roli.
\newblock Evasion attacks against machine learning at test time.
\newblock In {\em Joint European conference on machine learning and knowledge discovery in databases}, pages 387--402. Springer, 2013.

\bibitem{szegedy2014intriguing}
Christian Szegedy, Wojciech Zaremba, Ilya Sutskever, Joan Bruna, Dumitru Erhan, Ian Goodfellow, and Rob Fergus.
\newblock Intriguing properties of neural networks.
\newblock {\em arXiv preprint arXiv:1312.6199}, 2013.

\bibitem{papernot2017practical}
Nicolas Papernot, Patrick McDaniel, Ian Goodfellow, Somesh Jha, Z~Berkay Celik, and Ananthram Swami.
\newblock Practical black-box attacks against machine learning.
\newblock In {\em Proceedings of the 2017 ACM on Asia conference on computer and communications security}, pages 506--519, 2017.

\bibitem{lord2022attacking}
Nicholas~A. Lord, Romain Mueller, and Luca Bertinetto.
\newblock Attacking deep networks with surrogate-based adversarial black-box methods is easy.
\newblock 2022.

\bibitem{tashiro2020diversity}
Yusuke Tashiro, Yang Song, and Stefano Ermon.
\newblock Diversity can be transferred: Output diversification for white-and black-box attacks.
\newblock {\em Advances in neural information processing systems}, 33:4536--4548, 2020.

\bibitem{guo2021robust_collaborative}
Shangwei Guo, Xu~Zhang, Fei Yang, Tianwei Zhang, Yan Gan, Tao Xiang, and Yang Liu.
\newblock Robust and privacy-preserving collaborative learning: A comprehensive survey.
\newblock {\em arXiv preprint arXiv:2112.10183}, 2021.

\bibitem{security_imp_edge_comp}
Sina Ahmadi.
\newblock {Security Implications of Edge Computing in Cloud Networks}.
\newblock {\em {Journal of Computer and Communications}}, 12(02):26--46, 2024.

\bibitem{pgd_attack}
Aleksander Madry, Aleksandar Makelov, Ludwig Schmidt, Dimitris Tsipras, and Adrian Vladu.
\newblock Towards deep learning models resistant to adversarial attacks.
\newblock In {\em 6th International Conference on Learning Representations, {ICLR} 2018, Vancouver, BC, Canada, April 30 - May 3, 2018, Conference Track Proceedings}, 2018.

\bibitem{ray_framework_moritz}
Philipp Moritz, Robert Nishihara, Stephanie Wang, Alexey Tumanov, Richard Liaw, Eric Liang, Melih Elibol, Zongheng Yang, William Paul, Michael~I. Jordan, and Ion Stoica.
\newblock Ray: a distributed framework for emerging ai applications.
\newblock In {\em Proceedings of the 13th USENIX Conference on Operating Systems Design and Implementation}, OSDI'18, page 561–577, USA, 2018. USENIX Association.

\bibitem{deepspeed_rasley}
Jeff Rasley, Samyam Rajbhandari, Olatunji Ruwase, and Yuxiong He.
\newblock Deepspeed: System optimizations enable training deep learning models with over 100 billion parameters.
\newblock In {\em Proceedings of the 26th ACM SIGKDD International Conference on Knowledge Discovery \& Data Mining}, KDD '20, page 3505–3506, New York, NY, USA, 2020. Association for Computing Machinery.

\bibitem{Sasi2024_iotsec_survey}
Tinshu Sasi, Arash~Habibi Lashkari, Rongxing Lu, Pulei Xiong, and Shahrear Iqbal.
\newblock A comprehensive survey on iot attacks: Taxonomy, detection mechanisms and challenges.
\newblock {\em Journal of Information and Intelligence}, 2(6):455--513, 2024.

\bibitem{Wei_IoTSec_survey}
Zihan Wei, Qiang Wei, Yangyang Geng, and Yahui Yang.
\newblock A survey on iot security: Vulnerability detection and protection.
\newblock In {\em Proceedings of the 2024 International Conference on Artificial Intelligence of Things and Computing}, AITC '24, page 1–8, New York, NY, USA, 2025. Association for Computing Machinery.

\bibitem{He_collaborative_attacks_2021}
Zecheng He, Tianwei Zhang, and Ruby~B. Lee.
\newblock Attacking and protecting data privacy in edge–cloud collaborative inference systems.
\newblock {\em IEEE Internet of Things Journal}, 8(12):9706--9716, 2021.

\bibitem{ZHANG2020_ioTsec_botnet_survey}
Xiaolu Zhang, Oren Upton, Nicole~Lang Beebe, and Kim-Kwang~Raymond Choo.
\newblock Iot botnet forensics: A comprehensive digital forensic case study on mirai botnet servers.
\newblock {\em Forensic Science International: Digital Investigation}, 32:300926, 2020.

\bibitem{SAYAKKARA2019_side_channel_survey}
Asanka Sayakkara, Nhien-An Le-Khac, and Mark Scanlon.
\newblock A survey of electromagnetic side-channel attacks and discussion on their case-progressing potential for digital forensics.
\newblock {\em Digital Investigation}, 29:43--54, 2019.

\bibitem{lyu2022privacy}
Lingjuan Lyu, Han Yu, Xingjun Ma, Chen Chen, Lichao Sun, Jun Zhao, Qiang Yang, and S~Yu Philip.
\newblock Privacy and robustness in federated learning: Attacks and defenses.
\newblock {\em IEEE transactions on neural networks and learning systems}, 2022.

\bibitem{liu2020backdoor_collaborative_attacks}
Yang Liu, Zhihao Yi, and Tianjian Chen.
\newblock Backdoor attacks and defenses in feature-partitioned collaborative learning.
\newblock {\em arXiv preprint arXiv:2007.03608}, 2020.

\bibitem{Xie_distributed_backdoor_attacks_2020}
Chulin Xie, Keli Huang, Pin-Yu Chen, and Bo~Li.
\newblock Dba: Distributed backdoor attacks against federated learning.
\newblock In {\em International Conference on Learning Representations}, 2020.

\bibitem{rossolini2024edge}
Giulio Rossolini, Tommaso Baldi, Alessandro Biondi, and Giorgio Buttazzo.
\newblock Edge-only universal adversarial attacks in distributed learning.
\newblock {\em arXiv preprint arXiv:2411.10500}, 2024.

\bibitem{CarliniAttack2017}
Nicholas Carlini and David~A. Wagner.
\newblock Towards evaluating the robustness of neural networks.
\newblock In {\em 2017 {IEEE} Symposium on Security and Privacy, {SP} 2017, San Jose, CA, USA, May 22-26, 2017}, pages 39--57. {IEEE} Computer Society, 2017.

\bibitem{goodfellow2014explaining}
Ian~J. Goodfellow, Jonathon Shlens, and Christian Szegedy.
\newblock Explaining and harnessing adversarial examples.
\newblock In {\em 3rd International Conference on Learning Representations, {ICLR} 2015, San Diego, CA, USA, May 7-9, 2015, Conference Track Proceedings}, 2015.

\bibitem{brau2022minimal}
Fabio Brau, Giulio Rossolini, Alessandro Biondi, and Giorgio Buttazzo.
\newblock On the minimal adversarial perturbation for deep neural networks with provable estimation error.
\newblock {\em IEEE Transactions on Pattern Analysis and Machine Intelligence}, pages 1--15, 2022.

\bibitem{ilyas2018black}
Andrew Ilyas, Logan Engstrom, Anish Athalye, and Jessy Lin.
\newblock Black-box adversarial attacks with limited queries and information.
\newblock In {\em International conference on machine learning}, pages 2137--2146. PMLR, 2018.

\bibitem{ilyas2018prior}
Andrew Ilyas, Logan Engstrom, and Aleksander Madry.
\newblock Prior convictions: Black-box adversarial attacks with bandits and priors.
\newblock {\em arXiv preprint arXiv:1807.07978}, 2018.

\bibitem{chen2017zoo}
Pin-Yu Chen, Huan Zhang, Yash Sharma, Jinfeng Yi, and Cho-Jui Hsieh.
\newblock Zoo: Zeroth order optimization based black-box attacks to deep neural networks without training substitute models.
\newblock In {\em Proceedings of the 10th ACM workshop on artificial intelligence and security}, pages 15--26, 2017.

\bibitem{guo2019simple}
Chuan Guo, Jacob Gardner, Yurong You, Andrew~Gordon Wilson, and Kilian Weinberger.
\newblock Simple black-box adversarial attacks.
\newblock In {\em International Conference on Machine Learning}, pages 2484--2493. PMLR, 2019.

\bibitem{cheng2019improving}
Shuyu Cheng, Yinpeng Dong, Tianyu Pang, Hang Su, and Jun Zhu.
\newblock Improving black-box adversarial attacks with a transfer-based prior.
\newblock {\em Advances in neural information processing systems}, 32, 2019.

\bibitem{demontis2019adversarial}
Ambra Demontis, Marco Melis, Maura Pintor, Matthew Jagielski, Battista Biggio, Alina Oprea, Cristina Nita-Rotaru, and Fabio Roli.
\newblock Why do adversarial attacks transfer? explaining transferability of evasion and poisoning attacks.
\newblock In {\em 28th USENIX security symposium (USENIX security 19)}, pages 321--338, 2019.

\bibitem{gu2023survey_atrans}
Jindong Gu, Jia Xiaojun, Pau de~Jorge, Yu~Wenqain, Liu Xinwei, Avery Ma, Xun Yuan, Hu~Anjun, Ashkan Khakzar, Li~Zhijiang, Cao Xiaochun, and Torr Philip.
\newblock A survey on transferability of adversarial examples across deep neural networks.
\newblock {\em arXiv preprint arXiv:2310.17626}, 2023.

\bibitem{zhu2024_survey_atrans_black_box}
Yanfei Zhu, Yaochi Zhao, Zhuhua Hu, Tan Luo, and Like He.
\newblock A review of black-box adversarial attacks on image classification.
\newblock {\em Neurocomputing}, 610:128512, 2024.

\bibitem{CAI_NEURIPS2022_surrogate}
Zikui Cai, Chengyu Song, Srikanth Krishnamurthy, Amit Roy-Chowdhury, and Salman Asif.
\newblock Blackbox attacks via surrogate ensemble search.
\newblock In S.~Koyejo, S.~Mohamed, A.~Agarwal, D.~Belgrave, K.~Cho, and A.~Oh, editors, {\em Advances in Neural Information Processing Systems}, volume~35, pages 5348--5362. Curran Associates, Inc., 2022.

\bibitem{Qin_Xiong_Yi_Hsieh_2023_AAAI_surrogate}
Yunxiao Qin, Yuanhao Xiong, Jinfeng Yi, and Cho-Jui Hsieh.
\newblock Training meta-surrogate model for transferable adversarial attack.
\newblock {\em Proceedings of the AAAI Conference on Artificial Intelligence}, 37(8):9516--9524, Jun. 2023.

\bibitem{tramer2016_stealing}
Florian Tram{\`e}r, Fan Zhang, Ari Juels, Michael~K Reiter, and Thomas Ristenpart.
\newblock Stealing machine learning models via prediction $\{$APIs$\}$.
\newblock In {\em 25th USENIX security symposium (USENIX Security 16)}, pages 601--618, 2016.

\bibitem{oliynyk2023know_stealing}
Daryna Oliynyk, Rudolf Mayer, and Andreas Rauber.
\newblock I know what you trained last summer: A survey on stealing machine learning models and defences.
\newblock {\em ACM Computing Surveys}, 55(14s):1--41, 2023.

\bibitem{hinton2015distilling}
Geoffrey Hinton, Oriol Vinyals, and Jeff Dean.
\newblock Distilling the knowledge in a neural network.
\newblock {\em arXiv preprint arXiv:1503.02531}, 2015.

\bibitem{xu2024survey_distill}
Xiaohan Xu, Ming Li, Chongyang Tao, Tao Shen, Reynold Cheng, Jinyang Li, Can Xu, Dacheng Tao, and Tianyi Zhou.
\newblock A survey on knowledge distillation of large language models.
\newblock {\em arXiv preprint arXiv:2402.13116}, 2024.

\bibitem{canella2019transient}
Claudio Canella, Michael Schwarz, Lukas Giner, Daniel Lee, and Daniel Gruss.
\newblock A systematic evaluation of transient execution attacks and defenses.
\newblock In {\em Proceedings of the 28th USENIX Security Symposium (USENIX Security '19)}, 2019.

\bibitem{Ding2020service}
Chuntao Ding, Ao~Zhou, Yunxin Liu, Rong~N. Chang, Ching-Hsien Hsu, and Shangguang Wang.
\newblock A cloud-edge collaboration framework for cognitive service.
\newblock {\em IEEE Transactions on Cloud Computing}, 10(3):1489--1499, 2022.

\bibitem{cordts2016cityscapes}
Marius Cordts, Mohamed Omran, Sebastian Ramos, Timo Rehfeld, Markus Enzweiler, Rodrigo Benenson, Uwe Franke, Stefan Roth, and Bernt Schiele.
\newblock The cityscapes dataset for semantic urban scene understanding.
\newblock In {\em Proceedings of the IEEE conference on computer vision and pattern recognition}, 2016.

\bibitem{heo2019comprehensive}
Byeongho Heo, Jeesoo Kim, Sangdoo Yun, Hyojin Park, Nojun Kwak, and Jin~Young Choi.
\newblock A comprehensive overhaul of feature distillation.
\newblock In {\em Proceedings of the IEEE/CVF international conference on computer vision}, pages 1921--1930, 2019.

\bibitem{cifar10}
Alex Krizhevsky, Vinod Nair, and Geoffrey Hinton.
\newblock Cifar-10 (canadian institute for advanced research).

\bibitem{matsubara2022split}
Yoshitomo Matsubara, Marco Levorato, and Francesco Restuccia.
\newblock Split computing and early exiting for deep learning applications: Survey and research challenges.
\newblock {\em ACM Computing Surveys}, 55(5):1--30, 2022.

\bibitem{vgg_simonyan2014very}
Karen Simonyan and Andrew Zisserman.
\newblock Very deep convolutional networks for large-scale image recognition.
\newblock {\em arXiv preprint arXiv:1409.1556}, 2014.

\bibitem{he2016deep}
Kaiming He, Xiangyu Zhang, Shaoqing Ren, and Jian Sun.
\newblock Deep residual learning for image recognition.
\newblock In {\em Proceedings of the IEEE conference on computer vision and pattern recognition}, pages 770--778, 2016.

\bibitem{howard2017mobilenets}
Andrew~G Howard, Menglong Zhu, Bo~Chen, Dmitry Kalenichenko, Weijun Wang, Tobias Weyand, Marco Andreetto, and Hartwig Adam.
\newblock Mobilenets: Efficient convolutional neural networks for mobile vision applications.
\newblock {\em arXiv preprint arXiv:1704.04861}, 2017.

\bibitem{guo2019subspace}
Yiwen Guo, Ziang Yan, and Changshui Zhang.
\newblock Subspace attack: Exploiting promising subspaces for query-efficient black-box attacks.
\newblock {\em Advances in Neural Information Processing Systems}, 32, 2019.

\end{thebibliography}

\end{document}